\documentclass[10pt,twocolumn,letterpaper]{article}

\usepackage{cvpr}
\usepackage{times}
\usepackage{epsfig}
\usepackage{graphicx}
\usepackage{subcaption}
\usepackage{adjustbox}
\usepackage{amsmath}
\usepackage{amssymb}
\usepackage{bm}
\usepackage{booktabs}
\usepackage{multirow}
\usepackage{tablefootnote}
\usepackage[]{algorithm}
\usepackage{algorithmic}
\usepackage{color}
\usepackage{wrapfig}
\usepackage{soul}
\newtheorem{assumption}{Assumption}

\usepackage[pagebackref=true,breaklinks=true,letterpaper=true,colorlinks,bookmarks=false]{hyperref}
\newcommand{\norm}[1]{\left\lVert#1\right\rVert}
\cvprfinalcopy 

\def\cvprPaperID{7876} 

\ifcvprfinal\fi
\begin{document}

\title{Towards Efficient Model Compression via Learned Global Ranking}

\author{Ting-Wu Chi$\text{n}^1$, Ruizhou Din$\text{g}^1$, Cha Zhan$\text{g}^2$, Diana Marculesc$\text{u}^{13}$\\
Carnegie Mellon Universit$\text{y}^1$, Microsoft Cloud and A$\text{I}^2$, The University of Texas at Austi$\text{n}^3$\\
{\tt\small \{tingwuc, rding\}@andrew.cmu.edu, chazhang@mirosoft.com, dianam@utexas.edu}
}

\maketitle

\begin{abstract}
Pruning convolutional filters has demonstrated its effectiveness in compressing ConvNets. Prior art in filter pruning requires users to specify a target model complexity (\textit{e.g.}, model size or FLOP count) for the resulting architecture. However, determining a target model complexity can be difficult for optimizing various embodied AI applications such as autonomous robots, drones, and user-facing applications. First, both the accuracy and the speed of ConvNets can affect the performance of the application. Second, the performance of the application can be hard to assess without evaluating ConvNets during inference. As a consequence, finding a sweet-spot between the accuracy and speed via filter pruning, which needs to be done in a trial-and-error fashion, can be time-consuming. This work takes a first step toward making this process more efficient by altering the goal of model compression to producing \textbf{a set} of ConvNets with various accuracy and latency trade-offs instead of producing one ConvNet targeting some pre-defined latency constraint. To this end, we propose to learn a \textbf{global ranking} of the filters across different layers of the ConvNet, which is used to obtain a set of ConvNet architectures that have different accuracy/latency trade-offs by pruning the bottom-ranked filters. Our proposed algorithm, \textit{LeGR}, is shown to be $2\times$ to $3\times$ faster than prior work while having comparable or better performance when targeting seven pruned ResNet-56 with different accuracy/FLOPs profiles on the CIFAR-100 dataset. Additionally, we have evaluated LeGR on ImageNet and Bird-200 with ResNet-50 and MobileNetV2 to demonstrate its effectiveness. Code available at \url{https://github.com/cmu-enyac/LeGR}.

\end{abstract}

\section{Introduction}\label{sec:intro}
Building on top of the success of visual perception~\cite{ren2015faster,he2017mask,he2016deep}, natural language processing~\cite{dai2019transformer,devlin2018bert}, and speech recognition~\cite{chiu2018state,park2019specaugment} with deep learning, researchers have started to explore the possibility of embodied AI applications. In embodied AI, the goal is to enable agents to take actions based on perceptions in some environments~\cite{savva2019habitat}. We envision that next generation embodied AI systems will run on mobile devices such as autonomous robots and drones, where compute resources are limited and thus, will require model compression techniques for bringing such intelligent agents into our lives.

In particular, pruning the convolutional filters in ConvNets, also known as filter pruning, has shown to be an effective technique~\cite{ye2018rethinking,liu2017learning,wen2016learning,li2016pruning} for trading accuracy for inference speed improvements. The core idea of filter pruning is to find the least important filters to prune by minimizing the accuracy degradation and maximizing the speed improvement. State-of-the-art filter pruning methods~\cite{gordon2018morphnet,he2018amc,liu2017learning,zhou2019accelerate,peng2019collaborative,dai2018compressing} require a target model complexity of the whole ConvNet (\textit{e.g.}, total filter count, FLOP count\footnote{The number of floating-point operations to be computed for a ConvNet to carry out an inference.}, model size, inference latency, \textit{etc.}) to obtain a pruned network. However, deciding a target model complexity for optimizing embodied AI applications can be hard. For example, considering delivery with autonomous drones, both inference speed and precision of object detectors can affect the drone velocity~\cite{boroujerdian2018mavbench}, which in turn affects the inference speed and precision\footnote{Higher velocity requires faster computation and might cause accuracy degradation due to the blurring effect of the input video stream.}. For an user-facing autonomous robot that has to perform complicated tasks such as MovieQA~\cite{tapaswi2016movieqa}, VQA~\cite{antol2015vqa}, and room-to-room navigation~\cite{anderson2018vision}, both speed and accuracy of the visual perception module can affect the user experience. These aforementioned applications require many iterations of trial-and-error to find the optimal trade-off point between speed and accuracy of the ConvNets. 

\begin{figure}[t]
    \centering
    \includegraphics[width=1\linewidth]{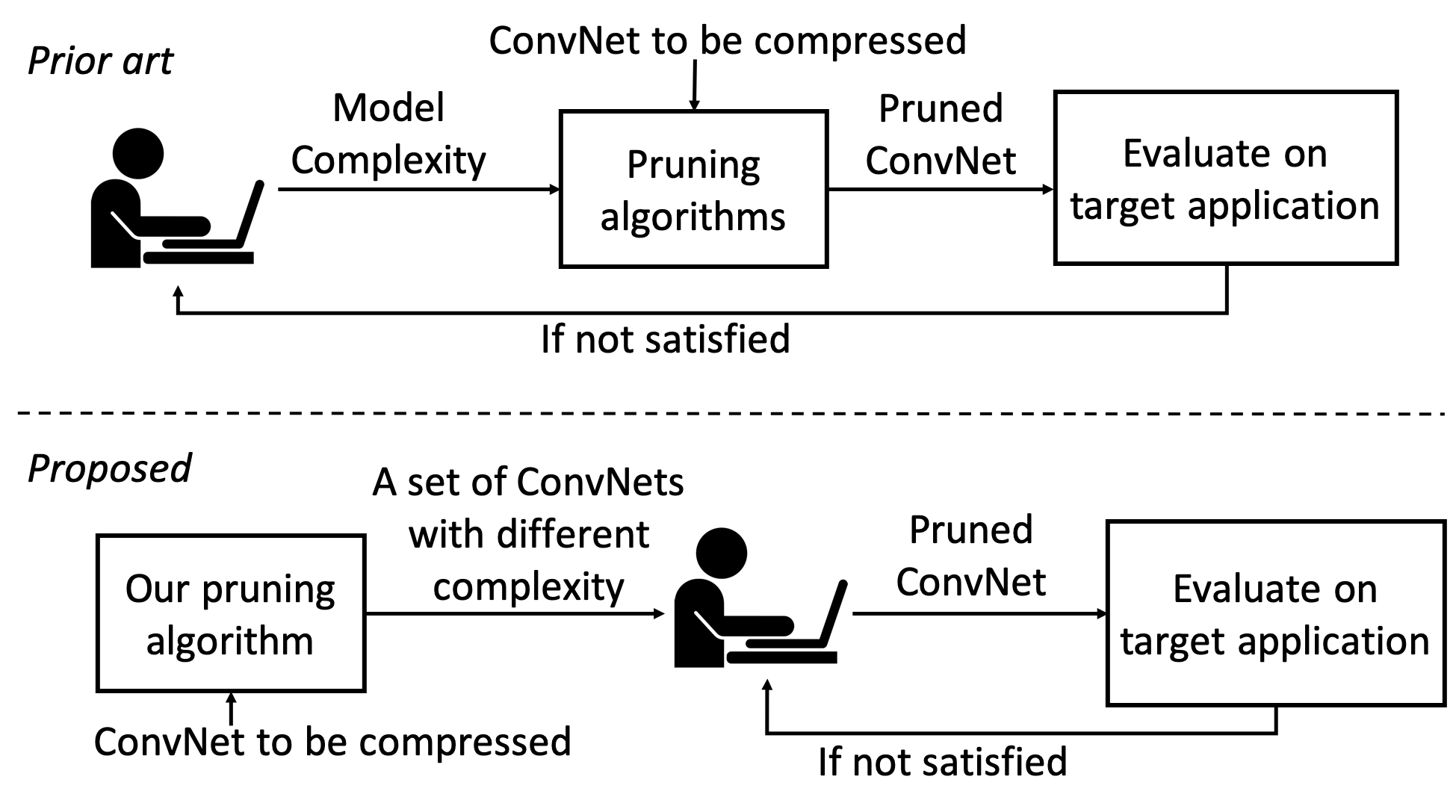}\\
    \caption{Using filter pruning to optimize ConvNets for embodied AI applications. Instead of producing one ConvNet for each pruning procedure as in prior art, our proposed method produces a set of ConvNets for practitioners to efficiently explore the trade-offs.}
    \label{fig:practical}
\end{figure}

More concretely, in these scenarios, practitioners would have to determine the sweet-spot for model complexity and accuracy in a trial-and-error fashion. Using an existing filter pruning algorithm many times to explore the impact of the different accuracy-vs.-speed trade-offs can be time-consuming. Figure~\ref{fig:practical} demonstrates the usage of filter pruning for optimizing ConvNets in aforementioned scenarios. With prior approaches, one has to go through the process of finding constraint-satisfying pruned-ConvNets via a pruning algorithm for every model complexity considered until practitioners are satisfied with the accuracy-vs.-speedup trade-off. Our work takes a first step toward alleviating the inefficiency in the aforementioned paradigm. We propose to alter the objective of pruning from outputting a single ConvNet with pre-defined model complexity to producing 
a set of ConvNets that have different accuracy/speed trade-offs, while achieving comparable accuracy with state-of-the-art methods (as shown in Figure~\ref{fig:pareto_imagenet}). In this fashion, the model compression overhead can be greatly reduced, which results in a more practical usage of filter pruning.

To this end, we propose \emph{learned global ranking (or LeGR)}, an algorithm that learns to rank convolutional filters across layers such that the ConvNet architectures of different speed/accuracy trade-offs can be obtained easily by dropping the bottom-ranked filters. The obtained architectures are then fine-tuned to generate the final models. In such a formulation, one can obtain a set of architectures by learning the ranking \textit{once}. We demonstrate the effectiveness of the proposed method with extensive empirical analyses using ResNet and MobileNetV2 on CIFAR-10/100, Bird-200, and ImageNet datasets. The main contributions of this work are as follows:

\begin{itemize}
    \item We propose learned global ranking (\emph{LeGR}), which produces a set of pruned ConvNets with different accuracy/speed trade-offs. \emph{LeGR} is shown to be faster than prior art in ConvNet pruning, while achieving comparable accuracy with state-of-the-art methods on three datasets and two types of ConvNets.
    \item Our formulation towards pruning is the first work that considers learning to rank filters across different layers globally, which addresses the limitation of prior art in magnitude-based filter pruning.
\end{itemize}
\section{Related Work}

Various methods have been developed to compress and/or accelerate ConvNets including weight quantization~\cite{rastegari2016xnor,zhu2016trained,Jacob_2018_CVPR,Jung_2019_CVPR,Yuan_2019_CVPR,hou2018lossaware,ding2019regularizing,choi2018bridging}, efficient convolution operators~\cite{howard2017mobilenets,he2019addressnet,wu2018shift,huang2017condensenet,zhao2019building}, neural architecture search~\cite{zhou2018resource,dai2018chamnet,cai2018proxylessnas,dong2018dpp,tan2018mnasnet,stamoulis2019single,stamoulis2018designing}, adjusting image resolution~\cite{pmlr-v97-tan19a,chin2019adascale}, and filter pruning, considered in this paper. Prior art on filter pruning can be grouped into two classes, depending on whether the architecture of the pruned-ConvNet is assumed to be given.

\paragraph{Pre-defined architecture} In this category, various work proposes different metrics to evaluate the importance of filters locally within each layer. For example, some prior work~\cite{li2016pruning,he2018soft} proposes to use $\ell_2$-norm of filter weights as the importance measure. On the other hand, other work has also investigated using the output discrepancy between the pruned and unpruned network as an importance measure~\cite{he2017channel,luo2017thinet}. However, the key drawback for methods that rank filters locally within a layer is that it is often hard to decide the overall target pruned architectures~\cite{he2018amc}. To cope with this difficulty, uniformly pruning the same portion of filters across all the layers is often adopted~\cite{he2018soft}.

\paragraph{Learned architecture} In this category, pruning algorithms learn the resulting structure automatically given a controllable parameter to determine the complexity of the pruned-ConvNet. To encourage weights with small magnitudes, Wen \emph{et al.}~\cite{wen2016learning} propose to add group-Lasso regularization to the filter norm to encourage filter weights to be zeros. Later, Liu \emph{et al.}~\cite{liu2017learning} propose to add Lasso regularization on the \emph{batch normalization layer} to achieve pruning during training.  Gordon \emph{et al.}~\cite{gordon2018morphnet} propose to add compute-weighted Lasso regularization on the filter norm. Huang \emph{et al.}~\cite{huang2018data} propose to add Lasso regularization on the output neurons instead of weights. While the regularization pushes unimportant filters to have smaller weights, the final thresholding applied globally assumes different layers to be equally important. Later, Louizos \emph{et al.}~\cite{louizos2018learning} have proposed $L_0$ regularization with stochastic relaxation. From a Bayesian perspective, Louizos \emph{et al.}~\cite{louizos2017bayesian} formulate pruning in a probabilistic fashion with a sparsity-induced prior. Similarly, Zhou \emph{et al.}~\cite{zhou2019accelerate} propose to model inter-layer dependency. From a different perspective, He \emph{et al.} propose an automated model compression framework (AMC)~\cite{he2018amc}, which uses reinforcement learning to search for a ConvNet that satisfies user-specified complexity constraints. 

While these prior approaches provide competitive pruned-ConvNets under a given target model complexity, it is often hard for one to specify the complexity parameter when compressing a ConvNet in embodied AI applications. To cope with this, our work proposes to generate a set of pruned-ConvNets across different complexity values rather than a single pruned-ConvNet under a target model complexity. 

We note that some prior work gradually prunes the ConvNet by alternating between pruning out a filter and fine-tuning, and thus, can also obtain a set of pruned-ConvNets with different complexities. For example, Molchanov \emph{et al.}~\cite{molchanov2016pruning} propose to use the normalized Taylor approximation of the loss as a measure to prune filters. Specifically, they greedily prune one filter at a time and fine-tune the network for a few gradient steps before the pruning proceeds. Following this paradigm, Theis \emph{et al.}~\cite{theis2018faster} propose to switch from first-order Taylor to Fisher information. However, our experiment results show that the pruned-ConvNet obtained by these methods have inferior accuracy compared to the methods that generate a single pruned ConvNet.

To obtain a set of ConvNets across different complexities with competitive performance, we propose to learn a global ranking of filters across different layers in a data-driven fashion such that architectures with different complexities can be obtained by pruning out the bottom-ranked filters. 

\section{Learned Global Ranking}\label{sec:LeGR-Lite}
The core idea of the proposed method is to learn a ranking for filters across different layers such that a ConvNet of a given complexity can be obtained easily by pruning out the bottom rank filters. In this section, we discuss our assumptions and formulation toward achieving this goal. 

As mentioned earlier in Section~\ref{sec:intro}, often both accuracy and latency of a ConvNet affect the performance of the overall application. The goal for model compression in these settings is to explore the accuracy-vs.-speed trade-off for finding a sweet-spot for a particular application using model compression. Thus, in this work, we use FLOP count for the model complexity to sample ConvNets. As we will show in Section~\ref{sec:latency}, we find FLOP count to be predictive for latency.

\subsection{Global Ranking}
To obtain pruned-ConvNets with different FLOP counts, we propose to learn the filter ranking globally across layers. In such a formulation, the global ranking for a given ConvNet just needs to be learned once and can be used to obtain ConvNets with different FLOP counts. However, there are two challenges for such a formulation. First, the global ranking formulation enforces an assumption that the top-performing smaller ConvNets are a proper subset of the top-performing larger ConvNets. The assumption might be strong because there are many ways to set the filter counts across different layers to achieve a given FLOP count, which implies that there are opportunities where the top-performing smaller network can have more filter counts in some layers but fewer filter counts in some other layers compared to a top-performing larger ConvNet. Nonetheless, this assumption enables the idea of global filter ranking, which can generate pruned ConvNets with different FLOP counts efficiently. In addition, the experiment results in Section~\ref{sec:zeta_hat} show that the pruned ConvNets under this assumption are competitive in terms of performance with the pruned ConvNets obtained without this assumption. We state the subset assumption more formally below.
\begin{assumption}[Subset Assumption]\label{as:1}
  For an optimal pruned ConvNet with FLOP count $f$, let $\mathcal{F}(f)_l$ be the filter count for layer $l$. The subset assumption states that $\mathcal{F}(f)_l~\leq~\mathcal{F}(f')_l~\forall~l$ if $f~\leq~f'$.
\end{assumption}

Another challenge for learning a global ranking is the hardness of the problem. Obtaining an optimal global ranking can be expensive, \emph{i.e.}, it requires $O(K\times K!)$ rounds of network fine-tuning, where $K$ is the number of filters. Thus, to make it tractable, we assume the filter norm is able to rank filters locally (intra-layer-wise) but not globally (inter-layer-wise).
\begin{assumption}[Norm Assumption]\label{as:2}
  $\ell_2$ norm can be used to compare the importance of a filter within each layer, but not across layers.
\end{assumption}
We note that the \emph{norm assumption} is adopted and empirically verified by prior art~\cite{li2016pruning,yang2018netadapt,he2018amc}. For filter norms to be compared across layers, we propose to learn layer-wise affine transformations over filter norms. Specifically, the importance of filter $i$ is defined as follows:
\begin{align}
    I_i=\alpha_{l(i)} \norm{\bm{\Theta_i}}_2^2 + \kappa_{l(i)}\label{eq:1},
\end{align}
where $l(i)$ is the layer index for the $i^{th}$ filter, $\norm{\cdot}_2$ denotes $\ell_2$ norms, $\bm{\Theta_i}$ denotes the weights for the $i^{th}$ filter, and $\bm{\alpha}\in \mathbb{R}^L$, $\bm{\kappa}\in \mathbb{R}^L$ are learnable parameters that represent layer-wise scale and shift values, and $L$ denotes the number of layers. We will detail in Section~\ref{sec:learn-legr} how $\bm{\alpha}$-$\bm{\kappa}$ pairs are learned so as to maximize overall accuracy. 

\begin{figure*}[h]
    \centering
    \includegraphics[width=1\linewidth]{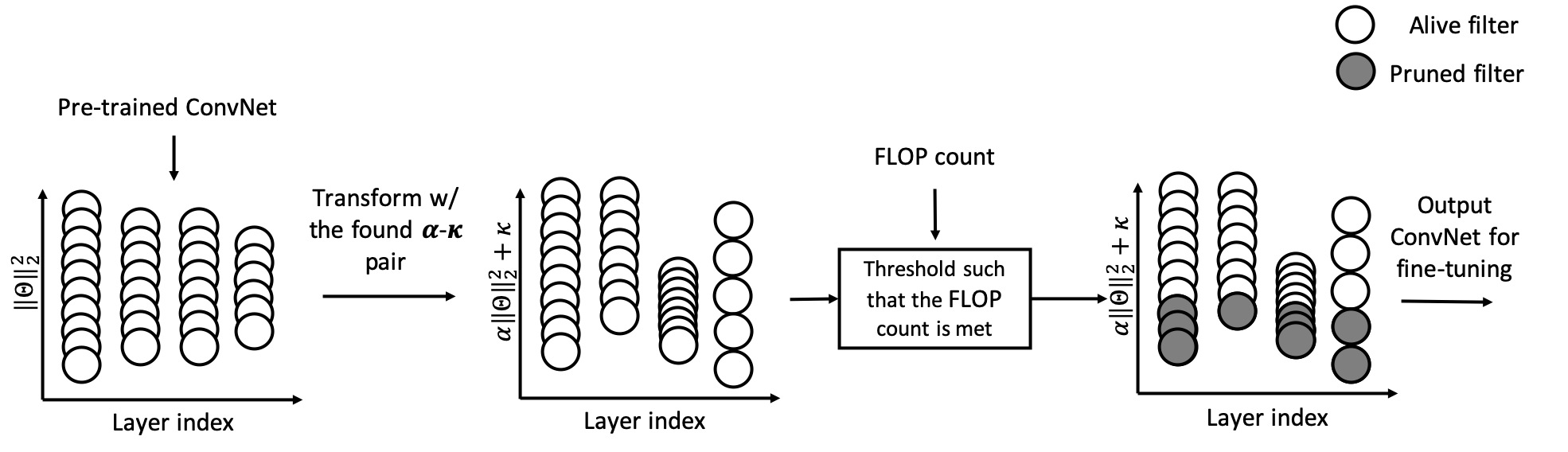}\\
    \caption{The flow of \emph{LeGR-Pruning}. $\norm{\Theta}_2^2$ represents the filter norm. Given the learned layer-wise affine transformations, \emph{i.e.}, the $\bm{\alpha}$-$\bm{\kappa}$ pair, \emph{LeGR-Pruning} returns filter masks that determine which filters are pruned. After \emph{LeGR-Pruning}, the pruned network will be fine-tuned to obtain the final network. }
    \label{fig:legrflow}
\end{figure*}

Based on these learned affine transformations from Eq.~(\ref{eq:1}) (\emph{i.e.}, the $\bm{\alpha}$-$\bm{\kappa}$ pair), the \emph{LeGR}-based pruning proceeds by ranking filters globally using $\bm{I}$ and prunes away bottom-ranked filters, \emph{i.e.}, smaller in $\bm{I}$, such that the FLOP count of interest is met, as shown in Figure~\ref{fig:legrflow}. This process can be done efficiently without the need of training data (since the knowledge of pruning is encoded in the $\bm{\alpha}$-$\bm{\kappa}$ pair).

\subsection{Learning Global Ranking}\label{sec:learn-legr}
To learn $\bm{\alpha}$ and $\bm{\kappa}$, one can consider constructing a ranking with $\bm{\alpha}$ and $\bm{\kappa}$ and then uniformly sampling ConvNets across different FLOP counts to evaluate the ranking. However, ConvNets obtained with different FLOP counts have drastically different validation accuracy, and one has to know the Pareto curve\footnote{A Pareto curve describes the optimal trade-off curve between two metrics of interest. Specifically, one cannot obtain improvement in one metric without degrading the other metric. The two metrics we considered in this work are accuracy and FLOP count.} of pruning to normalize the validation accuracy across ConvNets obtained with different FLOP counts. To address this difficulty, we propose to evaluate the validation accuracy of the ConvNet obtained from the lowest considered FLOP count as the objective for the ranking induced by the $\bm{\alpha}$-$\bm{\kappa}$ pair. Concretely, to learn $\bm{\alpha}$ and $\bm{\kappa}$, we treat \emph{LeGR} as an optimization problem:
\begin{align}
    \text{arg}\max_{\bm{\alpha},\bm{\kappa}}~Acc_{val}( \hat{\bm{\Theta_{l}}} )
\end{align}
where
\begin{align}
    \hat{\bm{\Theta_{l}}}=\text{LeGR-Pruning}(\bm{\alpha},\bm{\kappa}, \hat{\zeta}_{l}).
\end{align}
\emph{LeGR-Pruning} prunes away the bottom-ranked filters until the desired FLOP count is met as shown in Figure~\ref{fig:legrflow}. $\hat{\zeta}_{l}$ denotes the lowest FLOP count considered. As we will discuss later in Section~\ref{sec:zeta_hat}, we have also studied how $\hat{\zeta}$ affects the performance of the learned ranking, \emph{i.e.}, how the learned ranking affects the accuracy of the pruned networks.

\begin{algorithm}[h]
  \caption{Learning $\bm{\alpha},\bm{\kappa}$ with regularized EA}
  \label{alg:LeGR-Lite-full}
\begin{algorithmic}
    \STATE {\bfseries Input:} model $\bm{\Theta}$, lowest constraint $\hat{\zeta}_{l}$, random walk size $\sigma$, total search iterations $E$, sample size $S$, mutation ratio $u$, population size $P$, fine-tune iterations $\hat{\tau}$
    \STATE {\bfseries Output:} $\bm{\alpha},\bm{\kappa}$
    \STATE Initialize $Pool$ to a size $P$ queue
        
    \FOR{$e=1$ {\bfseries to} $E$}
        \STATE $\bm{\alpha}=\bm{1},~\bm{\kappa}=\bm{0}$
        \IF{$Pool$ has $S$ samples}
            \STATE $V$ = $Pool$.sample($S$)
            \STATE $\bm{\alpha},\bm{\kappa}$ = argmaxFitness($V$)
        \ENDIF
        \STATE $Layer$= Sample $u$\% layers to mutate
        \FOR{$l \in Layer$}
            \STATE std$_{l}$=computeStd([$M_i~\forall~i\in l$])
            \STATE $\bm{\alpha}_{l} = \bm{\alpha}_{l}\times \hat{\bm{\alpha}_{l}}$, where $\hat{\bm{\alpha}_{l}} \sim e^{\mathcal{N}(0,\sigma^2)}$
            \STATE $\bm{\kappa}_{l} = \bm{\kappa}_{l} + \hat{\bm{\kappa}_{l}}$, where $\hat{\bm{\kappa}_{l}} \sim \mathcal{N}(0,$std$_l)$
        \ENDFOR
        \STATE $\hat{\bm{\Theta_l}}$ = LeGR-Pruning-and-fine-tuning($\bm{\alpha},\bm{\kappa}$, $\hat{\zeta}_{l}$, $\hat{\tau}$, $\bm{\Theta}$)
        \STATE $Fitness = Acc_{val}(\hat{\bm{\Theta_l}})$
        \STATE $Pool$.replaceOldestWith($\bm{\alpha},\bm{\kappa},Fitness$)
    \ENDFOR
\end{algorithmic}
\end{algorithm}

Specifically, to learn the $\bm{\alpha}$-$\bm{\kappa}$ pair, we rely on approaches from hyper-parameter optimization literature. While there are several options for the optimization algorithm, we adopt the regularized evolutionary algorithm (EA) proposed in~\cite{real2018regularized} for its effectiveness in the neural architecture search space. The pseudo-code for our EA is outlined in Algorithm~\ref{alg:LeGR-Lite-full}. We have also investigated policy gradients for solving for the $\bm{\alpha}$-$\bm{\kappa}$ pair, which is shown in Appendix~\ref{sec:legrddpg}. We can equate each $\bm{\alpha}$-$\bm{\kappa}$ pair to a network architecture obtained by \emph{LeGR-Pruning}. Once a pruned architecture is obtained, we fine-tune the resulting architecture by $\hat{\tau}$ gradient steps and use its accuracy on the validation set\footnote{We split 10\% of the original training set to be used as validation set.} as the fitness (\emph{i.e.}, validation accuracy) for the corresponding $\bm{\alpha}$-$\bm{\kappa}$ pair. We note that we use $\hat{\tau}$ to approximate $\tau$ (fully fine-tuned steps) and we empirically find that $\hat{\tau}=200$ gradient updates work well under the pruning settings across the datasets and networks we study. More concretely, we first generate a pool of candidates ($\bm{\alpha}$ and $\bm{\kappa}$ values) and record the fitness for each candidate, and then repeat the following steps: (i) sample a subset from the candidates, (ii) identify the fittest candidate, (iii) generate a new candidate by mutating the fittest candidate and measure its fitness accordingly, and (iv) replace the oldest candidate in the pool with the generated one. To mutate the fittest candidate, we randomly select a subset of the layers $Layer$ and conduct one step of random-walk from their current values, \emph{i.e.}, $\alpha_l, \kappa_l~\forall~l\in Layer$.

We note that our layer-wise affine transformation formulation (Eq.~\ref{eq:1}) can be interpreted from an optimization perspective. That is, one can upper-bound the loss difference between a pre-trained ConvNet and its pruned-and-fine-tuned counterpart by assuming Lipschitz continuity on the loss function, as detailed in Appendix~\ref{appendix-optim}.
\section{Evaluations}\label{sec:eval}
\subsection{Datasets and Training Setting}
Our work is evaluated on various image classification benchmarks including CIFAR-10/100~\cite{krizhevsky2009learning}, ImageNet~\cite{russakovsky2015imagenet}, and Birds-200~\cite{wah2011caltech}. CIFAR-10/100 consists of 50k training images and 10k testing images with a total of 10/100 classes to be classified. ImageNet is a large scale image classification dataset that includes 1.2 million training images and 50k testing images with 1k classes to be classified. Also, we benchmark the proposed algorithm in a transfer learning setting since in practice, we want a small and fast model on some target datasets. Specifically, we use the Birds-200 dataset that consists of 6k training images and 5.7k testing images covering 200 bird species.

For Bird-200, we use 10\% of the training data as the validation set used for early stopping and to avoid over-fitting. The training scheme for CIFAR-10/100 follows~\cite{he2018soft}, which uses stochastic gradient descent with nesterov~\cite{nesterov1983method}, weight decay $5e^{-4}$, batch size 128, $1e^{-1}$ initial learning rate with decrease by 5$\times$ at epochs 60, 120, and 160, and train for 200 epochs in total. For control experiments with CIFAR-100 and Bird-200, the fine-tuning after pruning is done as follows: we keep all training hyper-parameters the same but change the initial learning rate to $1e^{-2}$ and train for 60 epochs (\emph{i.e.}, $\tau\approx 21$k). We drop the learning rate by $10\times$ at 30\%, 60\%, and 80\% of the total epochs, \emph{i.e.}, epochs 18, 36, and 48. To compare numbers with prior art on CIFAR-10 and ImageNet, we follow the number of iterations in~\cite{zhuang2018discrimination}. Specifically, for CIFAR-10 we fine-tuned for 400 epochs with initial learning rate $1e^{-2}$, drop by $5\times$ at epochs 120, 240, and 320. For ImageNet, we use pre-trained models and we fine-tuned the pruned models for 60 epochs with initial learning rate $1e^{-2}$, drop by $10\times$ at epochs 30 and 45.

For the hyper-parameters of \emph{LeGR}, we select $\hat{\tau}=200$, \emph{i.e.}, fine-tune for 200 gradient steps before measuring the validation accuracy when searching for the $\bm{\alpha}$-$\bm{\kappa}$ pair. We note that we do the same for \emph{AMC}~\cite{he2018amc} for a fair comparison. Moreover, we set the number of architectures explored to be the same with \emph{AMC}, \emph{i.e.}, 400. We set mutation rate $u=10$ and the hyper-parameter of the regularized evolutionary algorithm by following prior art~\cite{real2018regularized}. In the following experiments, we use the smallest $\zeta$ considered as $\hat{\zeta}_{l}$ to search for the learnable variables $\bm{\alpha}$ and $\bm{\kappa}$. The found $\bm{\alpha}$-$\bm{\kappa}$ pair is used to obtain the pruned networks at various FLOP counts. For example, for ResNet-56 with CIFAR-100 (Figure~\ref{fig:pareto_cifar100}), we use $\hat{\zeta}_{l}=20\%$ to obtain the $\bm{\alpha}$-$\bm{\kappa}$ pair and use the same $\bm{\alpha}$-$\bm{\kappa}$ pair to obtain the seven networks ($\zeta=20\%,...,80\%$) with the flow described in Figure~\ref{fig:legrflow}. The ablation of $\hat{\zeta}_{l}$ and $\hat{\tau}$ are detailed in Sec.~\ref{sec:ablation_tau}.

We prune filters across all the convolutional layers. We group dependent channels by summing up their importance measure and prune them jointly. The importance measure refers to the measure after learned affine transformations. Specifically, we group a channel in depth-wise convolution with its corresponding channel in the preceding layer. We also group channels that are summed together through residual connections.

\subsection{CIFAR-100 Results}\label{sec:main}
\begin{figure*}
    \centering
    \begin{subfigure}[t!]{0.6\linewidth}
        \centering
        \includegraphics[width=0.6\textwidth]{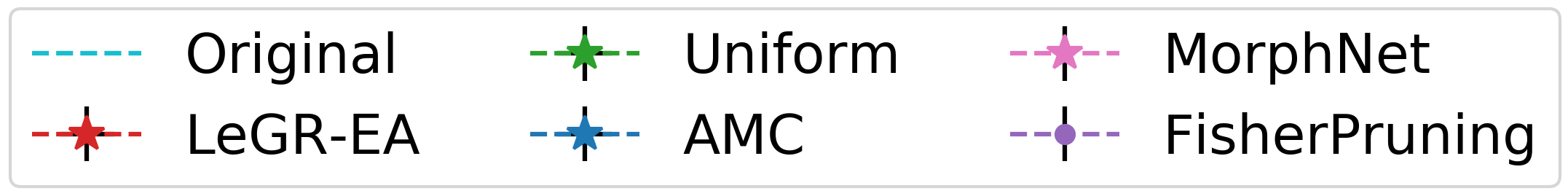}\\
        \includegraphics[width=0.47\textwidth]{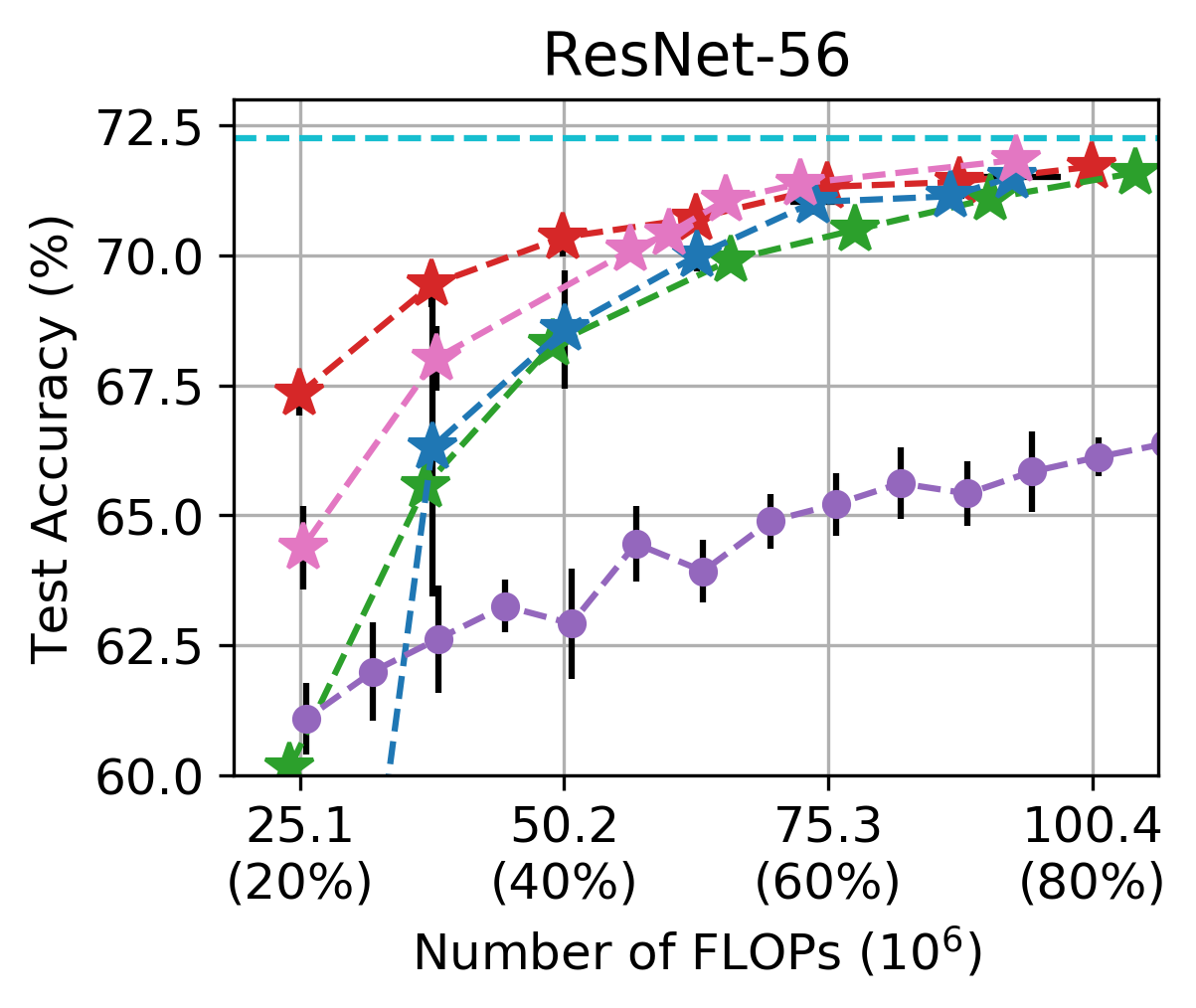}
        \includegraphics[width=0.47\textwidth]{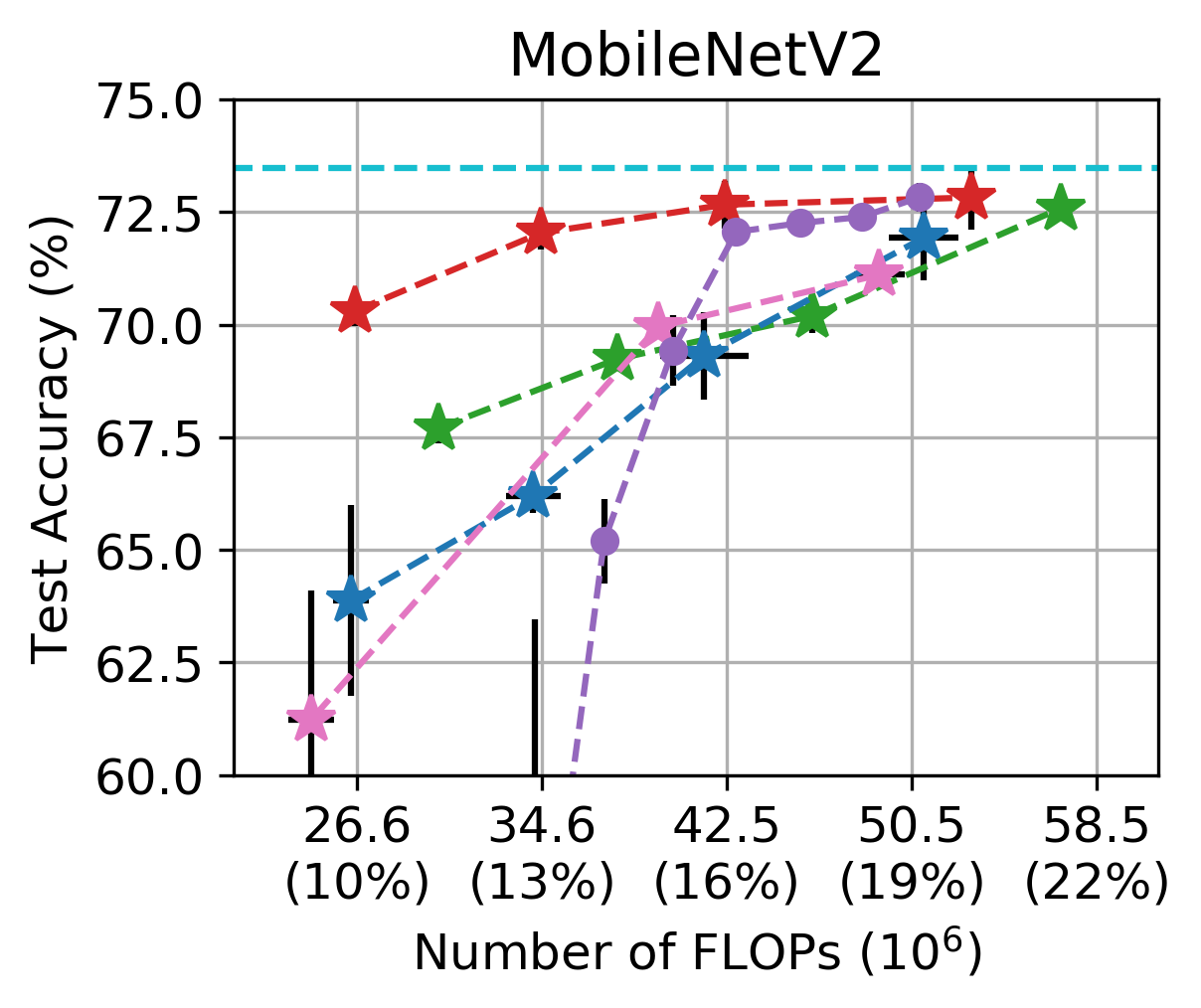}
        \caption{}
        \label{fig:pareto_cifar100}
    \end{subfigure}
    \begin{subfigure}[t!]{0.32\linewidth}
        \centering
        \includegraphics[width=1\textwidth]{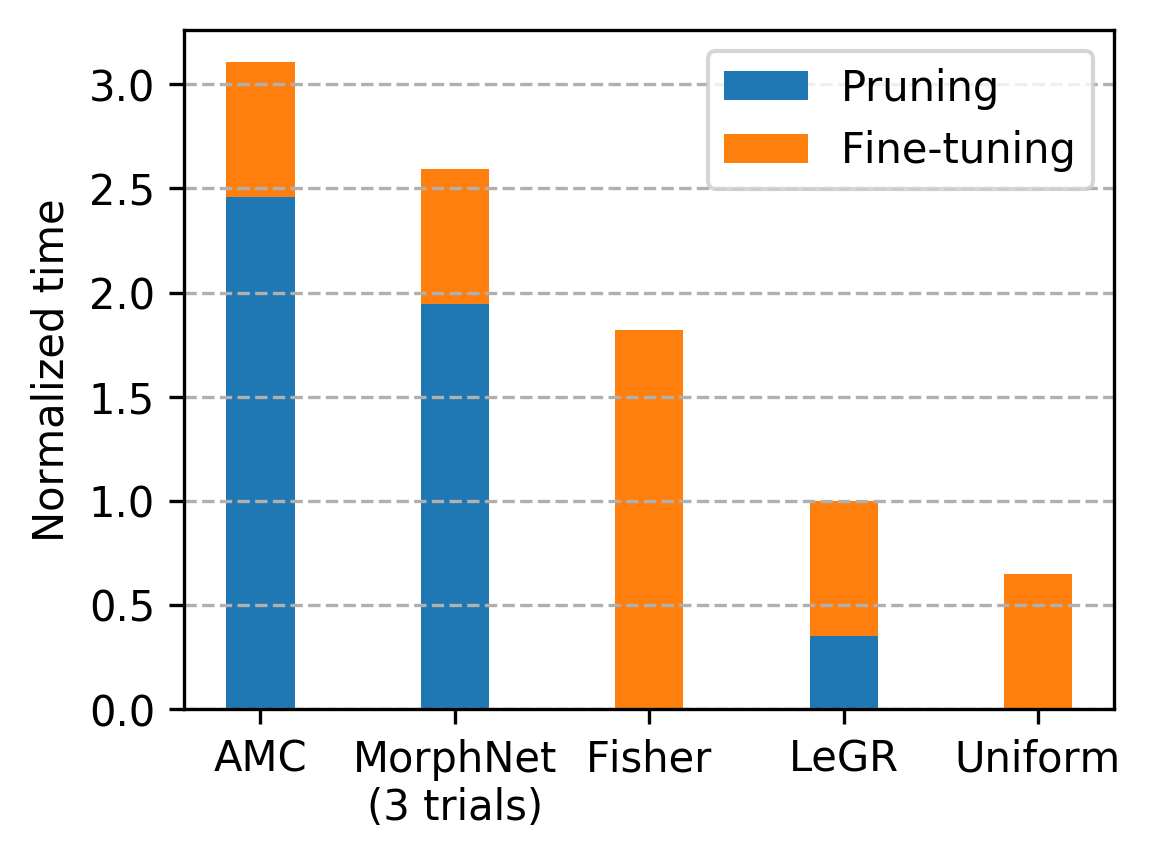}
        \caption{}
        \label{fig:pruning_cost}
    \end{subfigure}
    \caption{(a) The trade-off curve of pruning ResNet-56 and MobileNetV2 on CIFAR-100 using various methods. We average across three trials and plot the mean and standard deviation. (b) Training cost for seven ConvNets across FLOP counts using various methods targeting ResNet-56 on CIFAR-100. We report the average cost considering seven FLOP counts, \emph{i.e.}, 20\% to 80\% FLOP count in a step of 10\% on NVIDIA GTX 1080 Ti. The cost is normalized to the cost of LeGR.}
\end{figure*}

In this section, we consider ResNet-56 and MobileNetV2 and we compare \emph{LeGR} mainly with four filter pruning methods, \emph{i.e.}, \emph{MorphNet}~\cite{gordon2018morphnet}, \emph{AMC}~\cite{he2018amc}, \emph{FisherPruning}~\cite{theis2018faster}, and a baseline that prunes filters uniformly across layers. Specifically, the baselines are determined such that one dominant approach is selected from different groups of prior art. We select one approach~\cite{gordon2018morphnet} from pruning-while-learning approaches, one approach~\cite{he2018amc} from pruning-by-searching methods, one approach~\cite{theis2018faster} from continuous pruning methods, and a baseline extending magnitude-based pruning to various FLOP counts. We note that \emph{FisherPruning} is a continuous pruning method where we use 0.0025 learning rate and perform 500 gradient steps after each filter pruned following~\cite{theis2018faster}.

As shown in Figure~\ref{fig:pareto_cifar100}, we first observe that \emph{FisherPruning} does not work as well as other methods and we hypothesize the reason for it is that the small fixed learning rate in the fine-tuning phase makes it hard for the optimizer to get out of local optima. Additionally, we find that \emph{FisherPruning} prunes away almost all the filters for some layers. On the other hand, we find that all other approaches outperform the uniform baseline in a high-FLOP-count regime. However, both \emph{AMC} and \emph{MorphNet} have higher variances when pruned more aggressively. In both cases, \emph{LeGR} outperforms prior art, especially in the low-FLOP-count regime.

More importantly, our proposed method aims to alleviate the cost of pruning when the goal is to explore the trade-off curve between accuracy and inference latency. From this perspective, our approach outperforms prior art by a significant margin. More specifically, we measure the average time of each algorithm to obtain the seven pruned ResNet-56 across the FLOP counts in Figure~\ref{fig:pareto_cifar100} using our hardware (\emph{i.e.}, NVIDIA GTX 1080 Ti). Figure~\ref{fig:pruning_cost} shows the efficiency of \emph{AMC}, \emph{MorphNet}, \emph{FisherPruning}, and the proposed \emph{LeGR}. The cost can be broken down into two parts: (1) pruning: the time it takes to search for a network that has some pre-defined FLOP count and (2) fine-tuning: the time it takes for fine-tuning the weights of a pruned network. For \emph{MorphNet}, we consider three trials for each FLOP count to find an appropriate hyper-parameter $\lambda$ to meet the FLOP count of interest. The numbers are normalized to the cost of \emph{LeGR}. In terms of pruning time, \emph{LeGR} is $7\times$ and $5\times$ faster than \emph{AMC} and \emph{MorphNet}, respectively. The efficiency comes from the fact that \emph{LeGR} only searches the $\bm{\alpha}$-$\bm{\kappa}$ pair once and re-uses it across FLOP counts. In contrast, both \emph{AMC} and \emph{MorphNet} have to search for networks for every FLOP count considered. \emph{FisherPruning} always prune one filter at a time, and therefore the lowest FLOP count level considered determines the pruning time, regardless of how many FLOP count levels we are interested in.

\subsection{Comparison with Prior Art}
Although the goal of this work is to develop a model compression method that produces a set of ConvNets across different FLOP counts, we also compare our method with prior art that focuses on generating a ConvNet for a specified FLOP count.
\vspace{-10pt}
\paragraph{CIFAR-10} In Table~\ref{table:cifar10}, we compare \emph{LeGR} with prior art that reports results on CIFAR-10. First, for ResNet-56, we find that \emph{LeGR} outperforms most of the prior art in both FLOP count and accuracy dimensions and performs similarly to~\cite{he2018soft,zhuang2018discrimination}. For VGG-13, \emph{LeGR} achieves significantly better results compared to prior art.

\begin{table}[h]
\caption{Comparison with prior art on CIFAR-10. We group methods into sections according to different FLOP counts. Values for our approaches are averaged across three trials and we report the mean and standard deviation. We use boldface to denote the best numbers and use $^*$ to denote our implementation. The accuracy is represented in the format of \emph{pre-trained} $\mapsto$ \emph{pruned-and-fine-tuned}.}
\vskip 0.15in
\begin{center}
\begin{small}
\begin{sc}
\begin{adjustbox}{max width=1\linewidth}
\begin{tabular}{c|ccc}
\toprule
Network & Method & Acc. (\%) & MFLOP count\\
\midrule
\multirow{9}{*}{ResNet-56}&PF~\cite{li2016pruning} & 93.0 $\xrightarrow{}$ 93.0 & 90.9 (72\%)\\
&Taylor~\cite{molchanov2016pruning}$^*$ & \textbf{93.9} $\xrightarrow{}$ 93.2 & 90.8 (72\%)\\
&\textbf{LeGR} & \textbf{93.9} $\xrightarrow{}$ \textbf{94.1$\pm$0.0} & \textbf{87.8} (70\%)\\
\cline{2-4}
&DCP-Adapt~\cite{zhuang2018discrimination} & 93.8 $\xrightarrow{}$ \textbf{93.8} & 66.3 (53\%)\\
&CP~\cite{he2017channel} & 92.8 $\xrightarrow{}$ 91.8 & 62.7 (50\%)\\
&AMC~\cite{he2018amc} & 92.8 $\xrightarrow{}$ 91.9 & 62.7 (50\%)\\
&DCP~\cite{zhuang2018discrimination} & 93.8 $\xrightarrow{}$ 93.5 & 62.7 (50\%)\\
&SFP~\cite{he2018soft} & 93.6$\pm$0.6 $\xrightarrow{}$ \textbf{93.4$\pm$0.3} & 59.4 (47\%)\\
&\textbf{LeGR} & \textbf{93.9} $\xrightarrow{}$ \textbf{93.7$\pm$0.2} & \textbf{58.9} (47\%)\\
\midrule
\multirow{4}{*}{VGG-13}&BC-GNJ~\cite{louizos2017bayesian} & 91.9 $\xrightarrow{}$ 91.4 & 141.5 (45\%)\\
&BC-GHS~\cite{louizos2017bayesian} & 91.9 $\xrightarrow{}$ 91 & 121.9 (39\%)\\
&VIBNet~\cite{dai2018compressing} & 91.9 $\xrightarrow{}$ 91.5 & 70.6 (22\%)\\
&\textbf{LeGR} & 91.9 $\xrightarrow{}$ \textbf{92.4$\pm$0.2} & 70.3 (22\%)\\
\bottomrule
\end{tabular}\label{table:cifar10}
\end{adjustbox}
\end{sc}
\end{small}
\end{center}
\vskip -0.1in
\end{table}

\paragraph{ImageNet Results}\label{sec:imagenet}
\begin{figure*}
    \centering
    \includegraphics[width=0.45\textwidth]{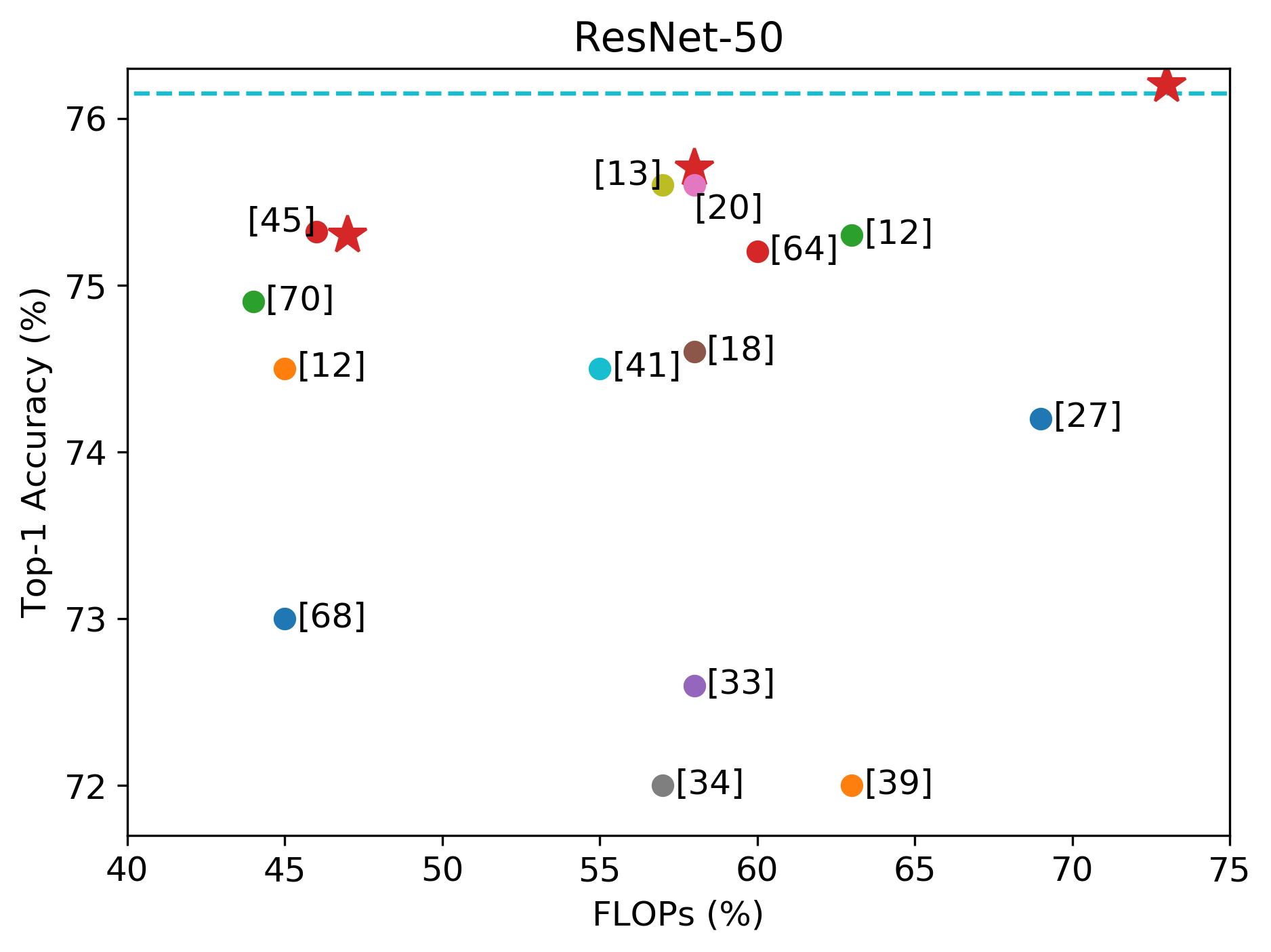}
    \includegraphics[width=0.45\textwidth]{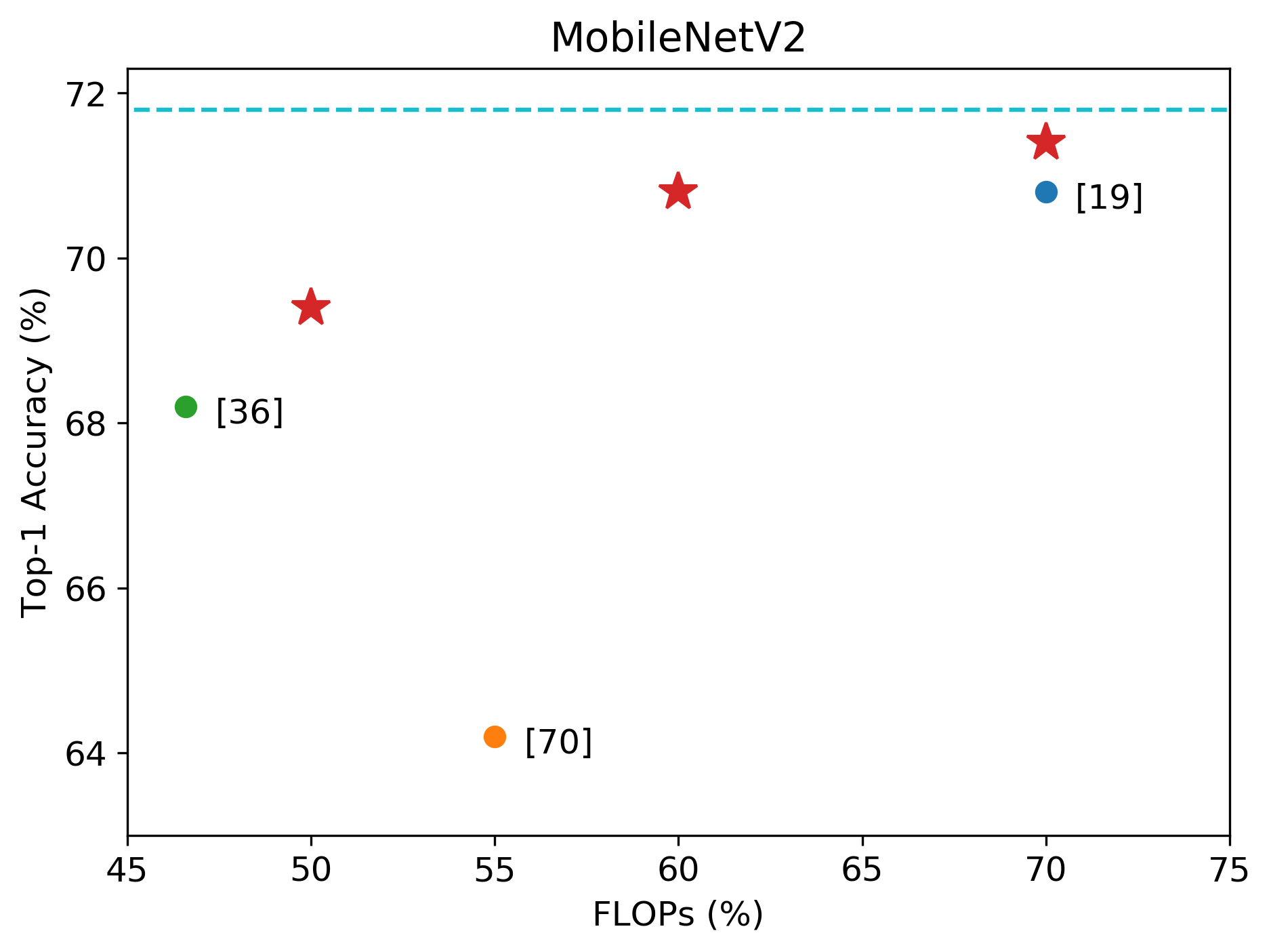}
    \\
    \includegraphics[width=0.7\textwidth]{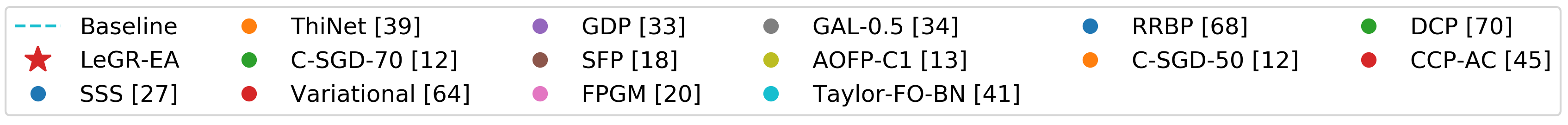}
    \includegraphics[width=0.23\textwidth]{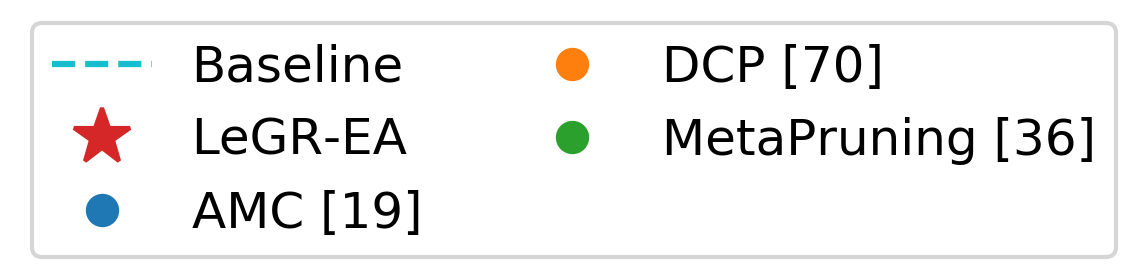}
    \caption{Results for ImageNet. LeGR is better or comparable compared to prior methods. Furthermore, its goal is to output a set of ConvNets instead of one ConvNet. The detailed numerical results are in Appendix~\ref{appendix-imagenet}.}\label{fig:pareto_imagenet}
\end{figure*}
For ImageNet, we prune ResNet-50 and MobileNetV2 with \emph{LeGR} to compare with prior art. For \emph{LeGR}, we learn the ranking using 47\% FLOP count for ResNet-50 and 50\% FLOP count for MobileNetV2, and use the learned ranking to obtain ConvNets for other FLOP counts of interest. We have compared to 17 prior methods that report pruning performance for ResNet-50 and/or MobileNetV2 on the ImageNet dataset. While our focus is on the fast exploration of the speed and accuracy trade-off curve for filter pruning, our proposed method is better or comparable compared to the state-of-the-art methods as shown in Figure~\ref{fig:pareto_imagenet}. The detailed numerical results are in Appendix~\ref{appendix-imagenet}. We would like to emphasize that to obtain a pruned-ConvNet with prior methods, one has to run the pruning algorithm for every FLOP count considered. In contrast, our proposed method learns the ranking once and uses it to obtain ConvNets across different FLOP counts.

\subsection{Transfer Learning: Bird-200}
\begin{figure}[h!]
    \vspace{15pt}
    \centering
    \includegraphics[width=0.8\linewidth]{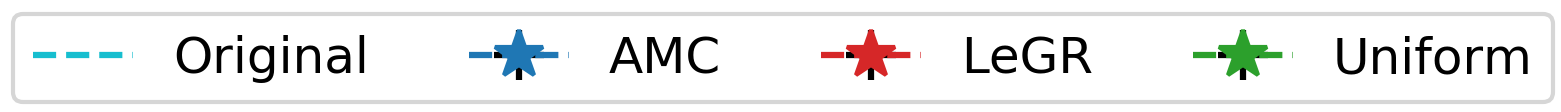}\\
    \includegraphics[width=0.47\linewidth]{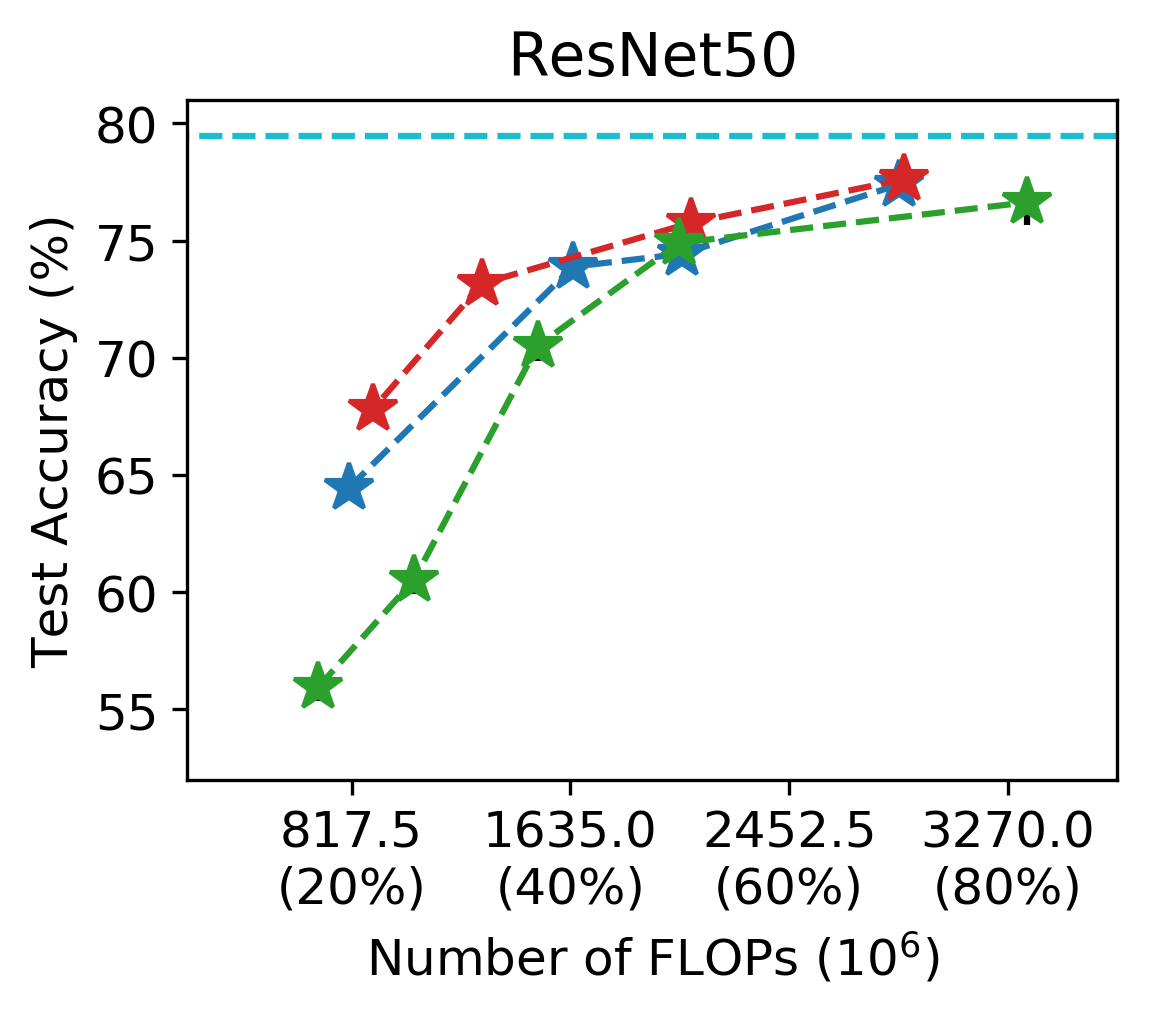}
    \includegraphics[width=0.47\linewidth]{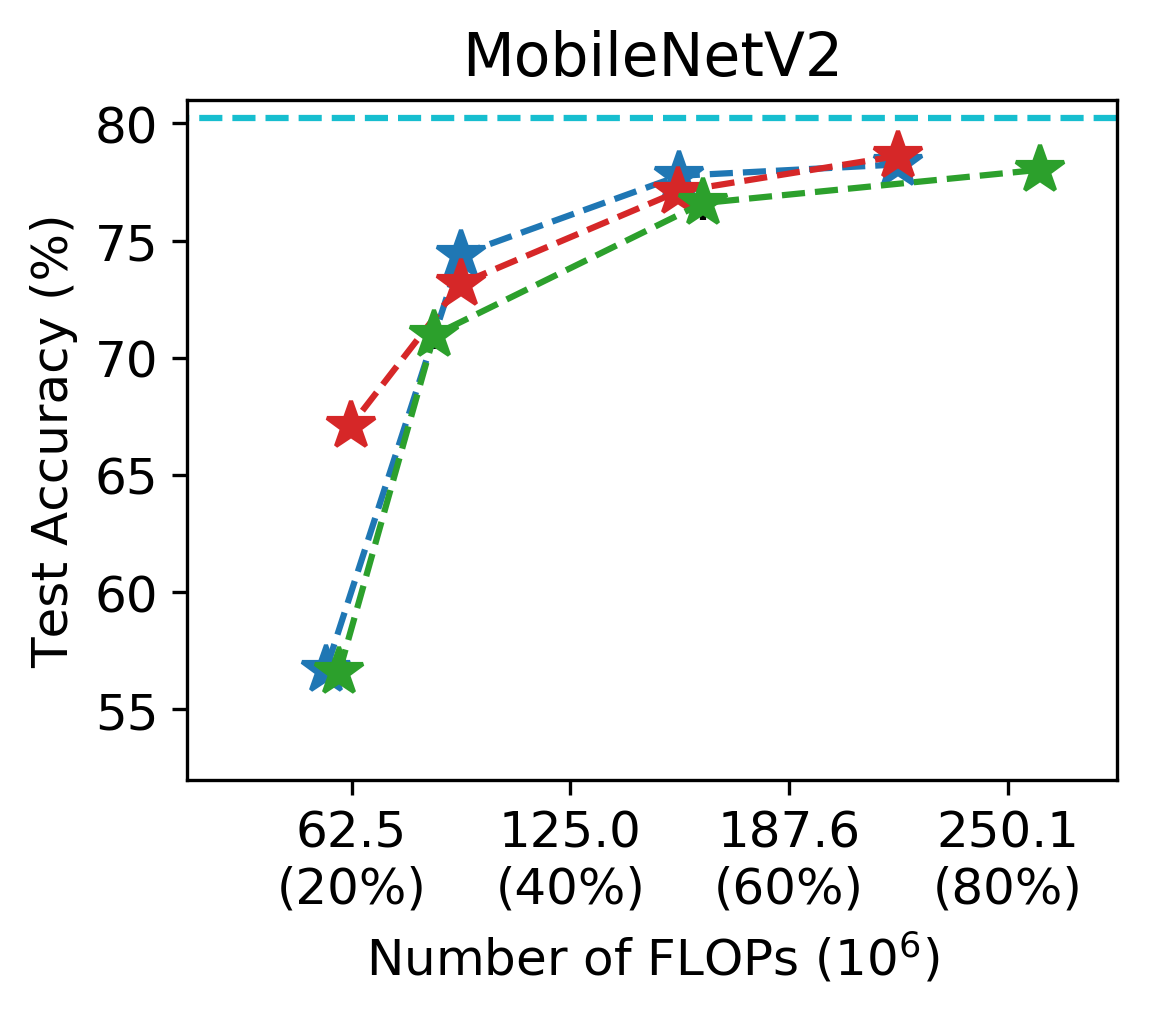}
    \caption{Results for Bird-200.}
    \label{fig:pareto_cub200}
\end{figure}

We analyze how \emph{LeGR} performs in a transfer learning setting where we have a model pre-trained on a large dataset, \emph{i.e.}, ImageNet, and we want to transfer its knowledge to adapt to a smaller dataset, \emph{i.e.}, Bird-200. We prune the fine-tuned network on the target dataset directly following the practice in prior art~\cite{zhong2018target,luo2017thinet}. We first obtain fine-tuned MobileNetV2 and ResNet-50 on the Bird-200 dataset with top-1 accuracy 80.2\% and 79.5\%, respectively. These are comparable to the reported values in prior art~\cite{li2018delta,Mallya_2018_CVPR}. As shown in Figure~\ref{fig:pareto_cub200}, we find that \emph{LeGR} outperforms \emph{Uniform} and \emph{AMC}, which is consistent with previous analyses in Section~\ref{sec:main}.

\vspace{-15pt}
\section{Ablation Study}
\subsection{Ranking Performance and $\hat{\zeta}_l$}\label{sec:zeta_hat}
\begin{figure}[h!]
    \centering
    \includegraphics[width=0.65\linewidth]{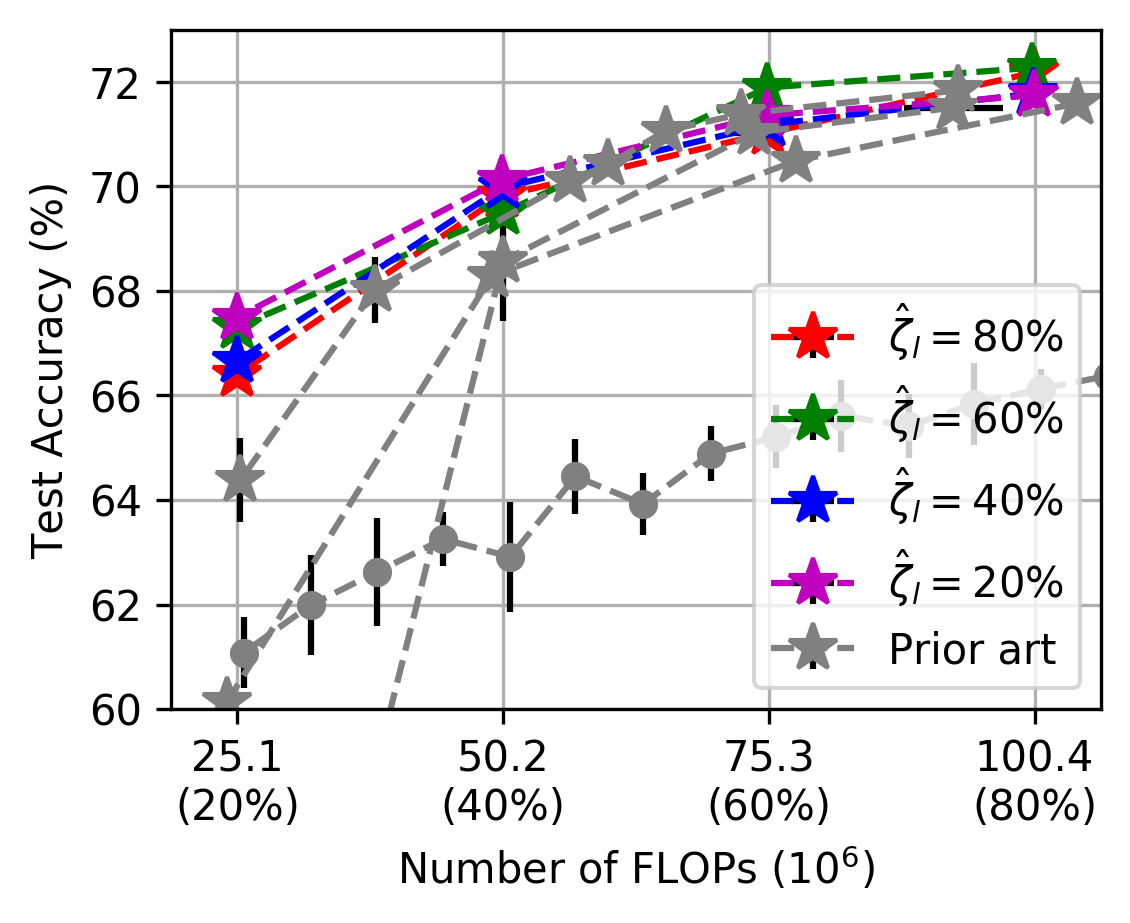}
    \caption{Robustness to the hyper-parameter $\hat{\zeta}_l$. Prior art is plotted as a reference (c.f. Figure~\ref{fig:pareto_cifar100}).}
    \label{fig:zeta-ablation}
\end{figure}
To learn the global ranking with \emph{LeGR} without knowing the Pareto curve in advance, we use the minimum considered FLOP count ($\hat{\zeta}_l$) during learning to evaluate the performance of a ranking. We are interested in understanding how this design choice affects the performance of \emph{LeGR}. Specifically, we try \emph{LeGR} targeting ResNet-56 for CIFAR-100 with $\hat{\zeta}_l~\in\{20\%,40\%,60\%,80\%\}$. As shown in Figure~\ref{fig:zeta-ablation}, we first observe that rankings learned using different FLOP counts have similar performances, which empirically supports Assumption~\ref{as:1}. More concretely, consider the network pruned to $40\%$ FLOP count by using the ranking learned at $40\%$ FLOP count. This case does not take advantage of the subset assumption because the entire learning process for learning $\bm{\alpha}$-$\bm{\kappa}$ is done only by looking at the performance of the $40\%$ FLOP count network. On the other hand, rankings learned using other FLOP counts but employed to obtain pruned-networks at $40\%$ FLOP count have exploited the subset assumption (\emph{e.g.}, the ranking learned for $80\%$ FLOP count can produce a competitive network for $40\%$ FLOP count). We find that \emph{LeGR} with or without employing Assumption~\ref{as:1} results in similar performance for the pruned networks.

\subsection{Fine-tuned Iterations}\label{sec:ablation_tau}
\begin{figure}[t]
    \centering
    \includegraphics[width=0.75\linewidth]{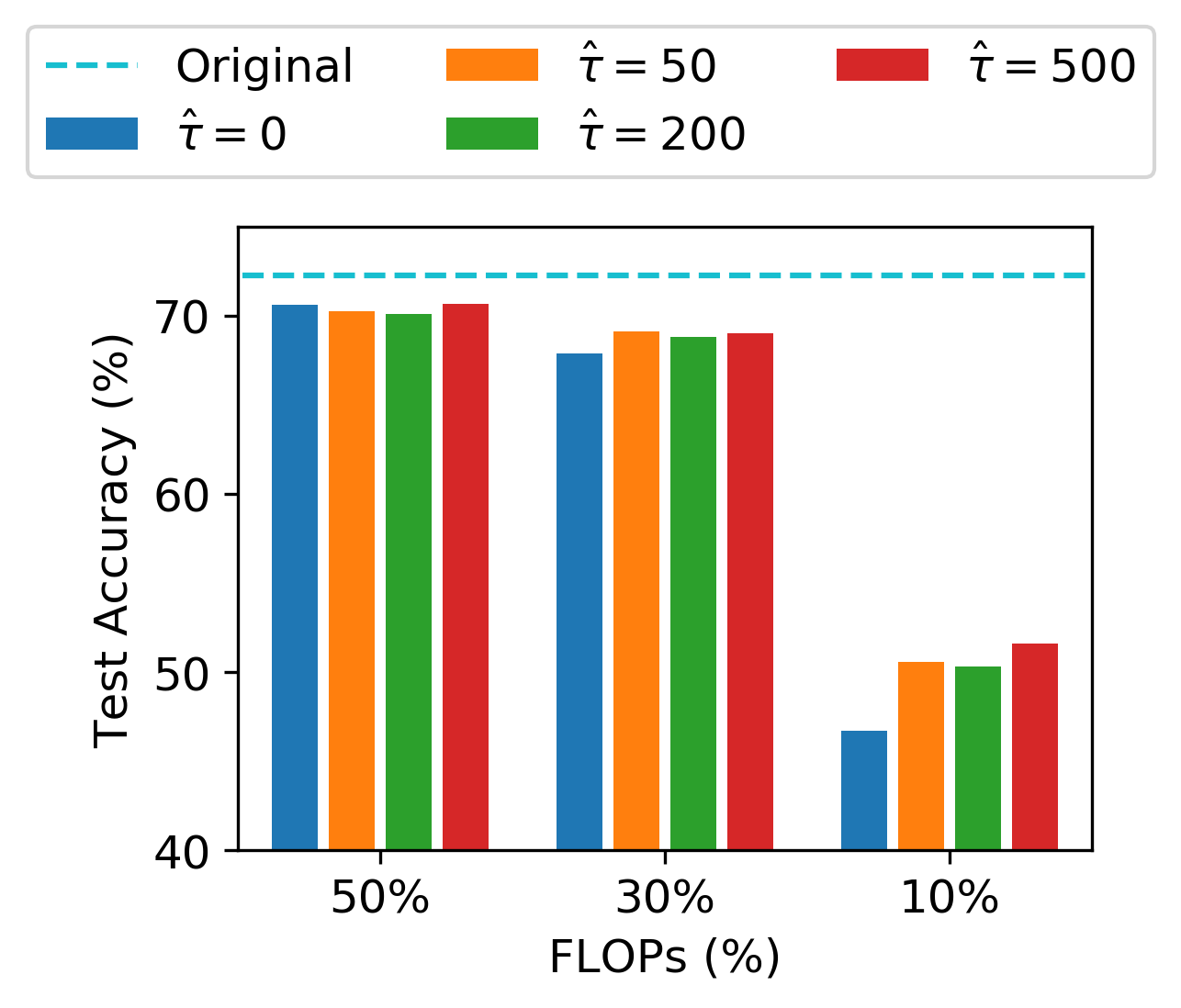}
    \caption{Pruning ResNet-56 for CIFAR-100 with \emph{LeGR} by learning $\bm{\alpha}$ and $\bm{\kappa}$ using different $\hat{\tau}$ and FLOP count constraints.}
    \label{fig:tau_analysis}
\end{figure}

Since we use $\hat{\tau}$ to approximate $\tau$ when learning the $\bm{\alpha}$-$\bm{\kappa}$ pair, it is expected that the closer $\hat{\tau}$ to $\tau$, the better the $\bm{\alpha}$-$\bm{\kappa}$ pair \emph{LeGR} can find. We use \emph{LeGR} to prune ResNet-56 for CIFAR-100 and learn $\bm{\alpha}$-$\bm{\kappa}$ at three FLOP counts $\hat{\zeta}_{l}\in\{10\%, 30\%, 50\%\}$. We consider $\zeta$ to be exactly $\hat{\zeta}_{l}$ in this case. For $\hat{\tau}$, we experiment with $\{0, 50, 200, 500\}$. We note that once the $\bm{\alpha}$-$\bm{\kappa}$ pair is learned, we use \emph{LeGR-Pruning} to obtain the pruned ConvNet, fine-tune it for $\tau$ steps, and plot the resulting test accuracy. In this experiment, $\tau$ is set to 21120 gradient steps (60 epochs). As shown in Figure~\ref{fig:tau_analysis}, the results align with our intuition in that there are diminishing returns in increasing $\hat{\tau}$. We observe that $\hat{\tau}$ affects the accuracy of the pruned ConvNets more when learning the ranking at a lower FLOP count level, which means in low-FLOP-count regimes, the validation accuracy after fine-tuning a few steps might not be representative. This makes sense since when pruning away a lot of filters, the network can be thought of as moving far away from the local optimal, where the gradient steps early in the fine-tuning phase are noisy. Thus, more gradient steps are needed before considering the accuracy to be representative of the fully-fine-tuned accuracy.

\subsection{FLOP count and Runtime}\label{sec:latency}
\begin{figure}[h]
    \centering
    \includegraphics[width=0.8\linewidth]{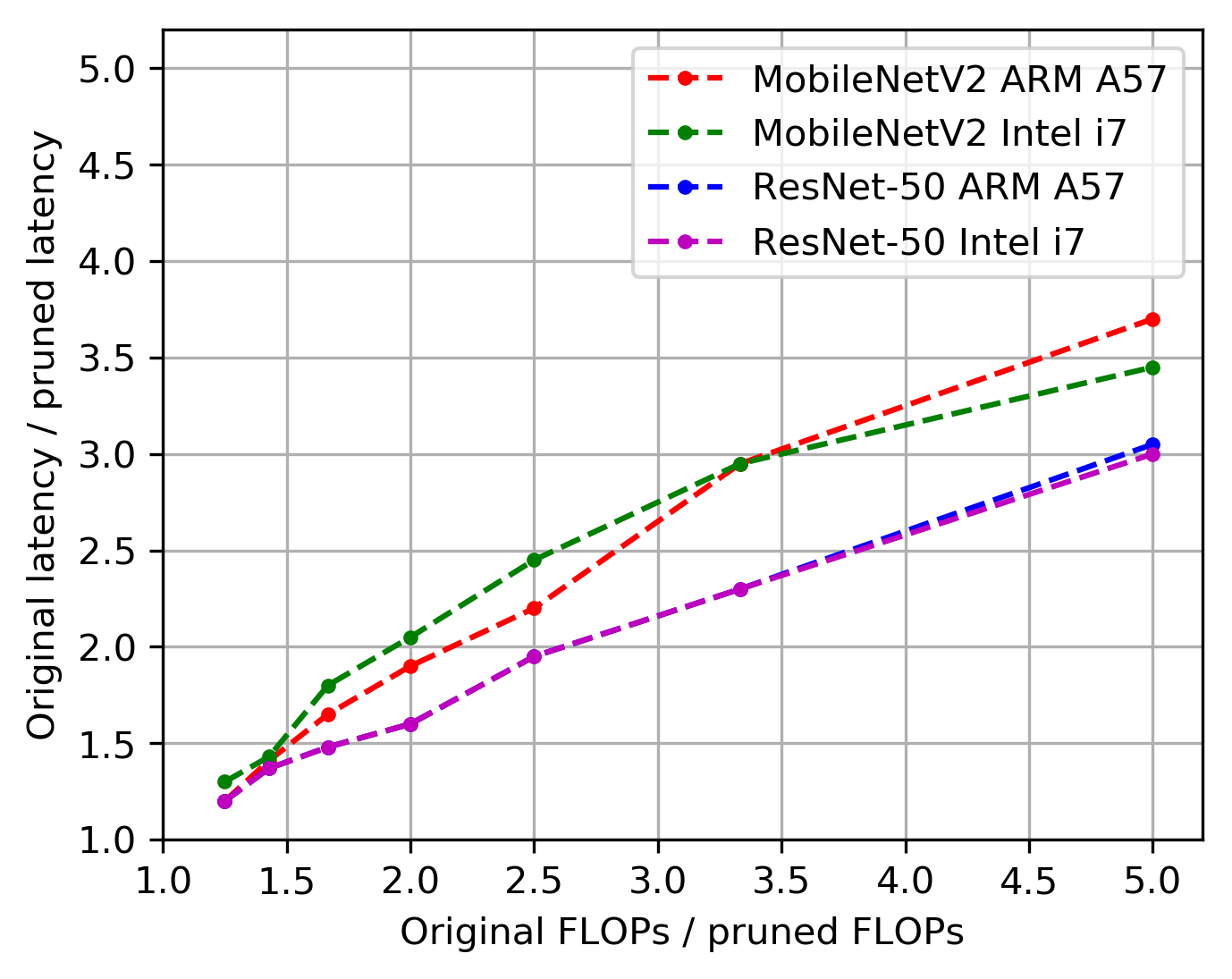}
    \caption{Latency reduction vs. FLOP count reduction. FLOP count reduction is indicative for latency reduction.}
    \label{fig:speedup}
\end{figure}
We demonstrate the effectiveness of filter pruning in wall-clock time speedup using ResNet-50 and MobileNetV2 on PyTorch 0.4 using two types of CPUs. Specifically, we consider both a desktop level CPU, \emph{i.e.}, Intel i7, and an embedded CPU, \emph{i.e.}, ARM A57, and use \emph{LeGR} as the pruning methodology. The input is a single RGB image of size 224x224 and the program (Python with PyTorch) is run using a single thread. As shown in Figure~\ref{fig:speedup}, filter pruning can produce near-linear acceleration (with a slope of approximately 0.6) without specialized software or hardware support.

\section{Conclusion}
To alleviate the bottleneck of using model compression in optimizing the ConvNets in a large system, we propose \emph{LeGR}, a novel formulation for practitioners to explore the accuracy-vs-speed trade-off efficiently via filter pruning. More specifically, we propose to learn layer-wise affine transformations over filter norms to construct a global ranking of filters. This formulation addresses the limitation that filter norms cannot be compared across layers in a learnable fashion and provides an efficient way for practitioners to obtain ConvNet architectures with different FLOP counts. Additionally, we provide a theoretical interpretation of the proposed affine transformation formulation. We conduct extensive empirical analyses using ResNet and MobileNetV2 on datasets including CIFAR, Bird-200, and ImageNet and show that \emph{LeGR} has less training cost to generate the pruned ConvNets across different FLOP counts compared to prior art while achieving comparable performance to state-of-the-art pruning methods.

\section*{Acknowledgement}
This research was supported in part by NSF CCF Grant No. 1815899, NSF CSR Grant No. 1815780, and NSF ACI Grant No. 1445606 at the Pittsburgh Supercomputing Center (PSC).

{\small
\bibliographystyle{ieee_fullname}
\bibliography{egbib}

\begin{thebibliography}{10}\itemsep=-1pt

\bibitem{anderson2018vision}
Peter Anderson, Qi Wu, Damien Teney, Jake Bruce, Mark Johnson, Niko
  S{\"u}nderhauf, Ian Reid, Stephen Gould, and Anton van~den Hengel.
\newblock Vision-and-language navigation: Interpreting visually-grounded
  navigation instructions in real environments.
\newblock In {\em Proceedings of the IEEE Conference on Computer Vision and
  Pattern Recognition}, pages 3674--3683, 2018.

\bibitem{antol2015vqa}
Stanislaw Antol, Aishwarya Agrawal, Jiasen Lu, Margaret Mitchell, Dhruv Batra,
  C Lawrence~Zitnick, and Devi Parikh.
\newblock Vqa: Visual question answering.
\newblock In {\em Proceedings of the IEEE international conference on computer
  vision}, pages 2425--2433, 2015.

\bibitem{boroujerdian2018mavbench}
Behzad Boroujerdian, Hasan Genc, Srivatsan Krishnan, Wenzhi Cui, Aleksandra
  Faust, and Vijay Reddi.
\newblock Mavbench: Micro aerial vehicle benchmarking.
\newblock In {\em 2018 51st Annual IEEE/ACM International Symposium on
  Microarchitecture (MICRO)}, pages 894--907. IEEE, 2018.

\bibitem{cai2018proxylessnas}
Han Cai, Ligeng Zhu, and Song Han.
\newblock Proxylessnas: Direct neural architecture search on target task and
  hardware.
\newblock {\em arXiv preprint arXiv:1812.00332}, 2018.

\bibitem{chin2019adascale}
Ting-Wu Chin, Ruizhou Ding, and Diana Marculescu.
\newblock Adascale: Towards real-time video object detection using adaptive
  scaling.
\newblock {\em arXiv preprint arXiv:1902.02910}, 2019.

\bibitem{chiu2018state}
Chung-Cheng Chiu, Tara~N Sainath, Yonghui Wu, Rohit Prabhavalkar, Patrick
  Nguyen, Zhifeng Chen, Anjuli Kannan, Ron~J Weiss, Kanishka Rao, Ekaterina
  Gonina, et~al.
\newblock State-of-the-art speech recognition with sequence-to-sequence models.
\newblock In {\em 2018 IEEE International Conference on Acoustics, Speech and
  Signal Processing (ICASSP)}, pages 4774--4778. IEEE, 2018.

\bibitem{choi2018bridging}
Jungwook Choi, Pierce I-Jen Chuang, Zhuo Wang, Swagath Venkataramani,
  Vijayalakshmi Srinivasan, and Kailash Gopalakrishnan.
\newblock Bridging the accuracy gap for 2-bit quantized neural networks (qnn).
\newblock {\em arXiv preprint arXiv:1807.06964}, 2018.

\bibitem{dai2018compressing}
Bin Dai, Chen Zhu, and David Wipf.
\newblock Compressing neural networks using the variational information
  bottleneck.
\newblock {\em arXiv preprint arXiv:1802.10399}, 2018.

\bibitem{dai2018chamnet}
Xiaoliang Dai, Peizhao Zhang, Bichen Wu, Hongxu Yin, Fei Sun, Yanghan Wang,
  Marat Dukhan, Yunqing Hu, Yiming Wu, Yangqing Jia, et~al.
\newblock Chamnet: Towards efficient network design through platform-aware
  model adaptation.
\newblock In {\em The IEEE Conference on Computer Vision and Pattern
  Recognition (CVPR)}, 2019.

\bibitem{dai2019transformer}
Zihang Dai, Zhilin Yang, Yiming Yang, William~W Cohen, Jaime Carbonell, Quoc~V
  Le, and Ruslan Salakhutdinov.
\newblock Transformer-xl: Attentive language models beyond a fixed-length
  context.
\newblock {\em arXiv preprint arXiv:1901.02860}, 2019.

\bibitem{devlin2018bert}
Jacob Devlin, Ming-Wei Chang, Kenton Lee, and Kristina Toutanova.
\newblock Bert: Pre-training of deep bidirectional transformers for language
  understanding.
\newblock {\em arXiv preprint arXiv:1810.04805}, 2018.

\bibitem{ding2019regularizing}
Ruizhou Ding, Ting-Wu Chin, Zeye Liu, and Diana Marculescu.
\newblock Regularizing activation distribution for training binarized deep
  networks.
\newblock 2019.

\bibitem{ding2019centripetal}
Xiaohan Ding, Guiguang Ding, Yuchen Guo, and Jungong Han.
\newblock Centripetal sgd for pruning very deep convolutional networks with
  complicated structure.
\newblock In {\em Proceedings of the IEEE Conference on Computer Vision and
  Pattern Recognition}, pages 4943--4953, 2019.

\bibitem{ding2019approximated}
Xiaohan Ding, Guiguang Ding, Yuchen Guo, Jungong Han, and Chenggang Yan.
\newblock Approximated oracle filter pruning for destructive cnn width
  optimization.
\newblock {\em arXiv preprint arXiv:1905.04748}, 2019.

\bibitem{dong2018dpp}
Jin-Dong Dong, An-Chieh Cheng, Da-Cheng Juan, Wei Wei, and Min Sun.
\newblock Dpp-net: Device-aware progressive search for pareto-optimal neural
  architectures.
\newblock {\em arXiv preprint arXiv:1806.08198}, 2018.

\bibitem{gordon2018morphnet}
Ariel Gordon, Elad Eban, Ofir Nachum, Bo Chen, Hao Wu, Tien-Ju Yang, and Edward
  Choi.
\newblock Morphnet: Fast \& simple resource-constrained structure learning of
  deep networks.
\newblock In {\em IEEE Conference on Computer Vision and Pattern Recognition
  (CVPR)}, 2018.

\bibitem{he2017mask}
Kaiming He, Georgia Gkioxari, Piotr Doll{\'a}r, and Ross Girshick.
\newblock Mask r-cnn.
\newblock In {\em Proceedings of the IEEE international conference on computer
  vision}, pages 2961--2969, 2017.

\bibitem{he2016deep}
Kaiming He, Xiangyu Zhang, Shaoqing Ren, and Jian Sun.
\newblock Deep residual learning for image recognition.
\newblock In {\em Proceedings of the IEEE conference on computer vision and
  pattern recognition}, pages 770--778, 2016.

\bibitem{he2018soft}
Yang He, Guoliang Kang, Xuanyi Dong, Yanwei Fu, and Yi Yang.
\newblock Soft filter pruning for accelerating deep convolutional neural
  networks.
\newblock In {\em IJCAI}, pages 2234--2240, 2018.

\bibitem{he2018amc}
Yihui He, Ji Lin, Zhijian Liu, Hanrui Wang, Li-Jia Li, and Song Han.
\newblock Amc: Automl for model compression and acceleration on mobile devices.
\newblock {\em arXiv preprint arXiv:1802.03494}, 2018.

\bibitem{he2019filter}
Yang He, Ping Liu, Ziwei Wang, Zhilan Hu, and Yi Yang.
\newblock Filter pruning via geometric median for deep convolutional neural
  networks acceleration.
\newblock In {\em Proceedings of the IEEE Conference on Computer Vision and
  Pattern Recognition}, pages 4340--4349, 2019.

\bibitem{he2019addressnet}
Yihui He, Xianggen Liu, Huasong Zhong, and Yuchun Ma.
\newblock Addressnet: Shift-based primitives for efficient convolutional neural
  networks.
\newblock In {\em 2019 IEEE Winter Conference on Applications of Computer
  Vision (WACV)}, pages 1213--1222. IEEE, 2019.

\bibitem{he2017channel}
Yihui He, Xiangyu Zhang, and Jian Sun.
\newblock Channel pruning for accelerating very deep neural networks.
\newblock In {\em International Conference on Computer Vision (ICCV)},
  volume~2, 2017.

\bibitem{hou2018lossaware}
Lu Hou and James~T. Kwok.
\newblock Loss-aware weight quantization of deep networks.
\newblock In {\em International Conference on Learning Representations}, 2018.

\bibitem{howard2017mobilenets}
Andrew~G Howard, Menglong Zhu, Bo Chen, Dmitry Kalenichenko, Weijun Wang,
  Tobias Weyand, Marco Andreetto, and Hartwig Adam.
\newblock Mobilenets: Efficient convolutional neural networks for mobile vision
  applications.
\newblock {\em arXiv preprint arXiv:1704.04861}, 2017.

\bibitem{huang2017condensenet}
Gao Huang, Shichen Liu, Laurens van~der Maaten, and Kilian~Q Weinberger.
\newblock Condensenet: An efficient densenet using learned group convolutions.
\newblock {\em group}, 3(12):11, 2017.

\bibitem{huang2018data}
Zehao Huang and Naiyan Wang.
\newblock Data-driven sparse structure selection for deep neural networks.
\newblock In {\em Proceedings of the European Conference on Computer Vision
  (ECCV)}, pages 304--320, 2018.

\bibitem{Huang_2018_ECCV}
Zehao Huang and Naiyan Wang.
\newblock Data-driven sparse structure selection for deep neural networks.
\newblock In {\em The European Conference on Computer Vision (ECCV)}, September
  2018.

\bibitem{Jacob_2018_CVPR}
Benoit Jacob, Skirmantas Kligys, Bo Chen, Menglong Zhu, Matthew Tang, Andrew
  Howard, Hartwig Adam, and Dmitry Kalenichenko.
\newblock Quantization and training of neural networks for efficient
  integer-arithmetic-only inference.
\newblock In {\em The IEEE Conference on Computer Vision and Pattern
  Recognition (CVPR)}, June 2018.

\bibitem{Jung_2019_CVPR}
Sangil Jung, Changyong Son, Seohyung Lee, Jinwoo Son, Jae-Joon Han, Youngjun
  Kwak, Sung~Ju Hwang, and Changkyu Choi.
\newblock Learning to quantize deep networks by optimizing quantization
  intervals with task loss.
\newblock In {\em The IEEE Conference on Computer Vision and Pattern
  Recognition (CVPR)}, June 2019.

\bibitem{krizhevsky2009learning}
Alex Krizhevsky and Geoffrey Hinton.
\newblock Learning multiple layers of features from tiny images.
\newblock Technical report, Citeseer, 2009.

\bibitem{li2016pruning}
Hao Li, Asim Kadav, Igor Durdanovic, Hanan Samet, and Hans~Peter Graf.
\newblock Pruning filters for efficient convnets.
\newblock {\em International Conference on Learning Representation (ICLR)},
  2017.

\bibitem{li2018delta}
Xingjian Li, Haoyi Xiong, Hanchao Wang, Yuxuan Rao, Liping Liu, and Jun Huan.
\newblock {DELTA}: {DEEP} {LEARNING} {TRANSFER} {USING} {FEATURE} {MAP} {WITH}
  {ATTENTION} {FOR} {CONVOLUTIONAL} {NETWORKS}.
\newblock In {\em International Conference on Learning Representations}, 2019.

\bibitem{lin2018accelerating}
Shaohui Lin, Rongrong Ji, Yuchao Li, Yongjian Wu, Feiyue Huang, and Baochang
  Zhang.
\newblock Accelerating convolutional networks via global \& dynamic filter
  pruning.
\newblock In {\em IJCAI}, pages 2425--2432, 2018.

\bibitem{lin2019towards}
Shaohui Lin, Rongrong Ji, Chenqian Yan, Baochang Zhang, Liujuan Cao, Qixiang
  Ye, Feiyue Huang, and David Doermann.
\newblock Towards optimal structured cnn pruning via generative adversarial
  learning.
\newblock In {\em Proceedings of the IEEE Conference on Computer Vision and
  Pattern Recognition}, pages 2790--2799, 2019.

\bibitem{liu2017learning}
Zhuang Liu, Jianguo Li, Zhiqiang Shen, Gao Huang, Shoumeng Yan, and Changshui
  Zhang.
\newblock Learning efficient convolutional networks through network slimming.
\newblock In {\em Computer Vision (ICCV), 2017 IEEE International Conference
  on}, pages 2755--2763. IEEE, 2017.

\bibitem{liu2019metapruning}
Zechun Liu, Haoyuan Mu, Xiangyu Zhang, Zichao Guo, Xin Yang, Tim Kwang-Ting
  Cheng, and Jian Sun.
\newblock Metapruning: Meta learning for automatic neural network channel
  pruning.
\newblock In {\em Proceedings of the IEEE International Conference on Computer
  Vision}, 2019.

\bibitem{louizos2017bayesian}
Christos Louizos, Karen Ullrich, and Max Welling.
\newblock Bayesian compression for deep learning.
\newblock In {\em Advances in Neural Information Processing Systems}, pages
  3288--3298, 2017.

\bibitem{louizos2018learning}
Christos Louizos, Max Welling, and Diederik~P. Kingma.
\newblock Learning sparse neural networks through $l_0$ regularization.
\newblock In {\em International Conference on Learning Representations}, 2018.

\bibitem{luo2017thinet}
Jian-Hao Luo, Jianxin Wu, and Weiyao Lin.
\newblock Thinet: A filter level pruning method for deep neural network
  compression.
\newblock {\em arXiv preprint arXiv:1707.06342}, 2017.

\bibitem{Mallya_2018_CVPR}
Arun Mallya and Svetlana Lazebnik.
\newblock Packnet: Adding multiple tasks to a single network by iterative
  pruning.
\newblock In {\em The IEEE Conference on Computer Vision and Pattern
  Recognition (CVPR)}, June 2018.

\bibitem{molchanov2019importance}
Pavlo Molchanov, Arun Mallya, Stephen Tyree, Iuri Frosio, and Jan Kautz.
\newblock Importance estimation for neural network pruning.
\newblock In {\em Proceedings of the IEEE Conference on Computer Vision and
  Pattern Recognition}, pages 11264--11272, 2019.

\bibitem{molchanov2016pruning}
Pavlo Molchanov, Stephen Tyree, Tero Karras, Timo Aila, and Jan Kautz.
\newblock Pruning convolutional neural networks for resource efficient
  inference.
\newblock {\em International Conference on Learning Representation (ICLR)},
  2017.

\bibitem{nesterov1983method}
Yurii~E Nesterov.
\newblock A method for solving the convex programming problem with convergence
  rate o (1/k\^{} 2).
\newblock In {\em Dokl. Akad. Nauk SSSR}, volume 269, pages 543--547, 1983.

\bibitem{park2019specaugment}
Daniel~S Park, William Chan, Yu Zhang, Chung-Cheng Chiu, Barret Zoph, Ekin~D
  Cubuk, and Quoc~V Le.
\newblock Specaugment: A simple data augmentation method for automatic speech
  recognition.
\newblock {\em arXiv preprint arXiv:1904.08779}, 2019.

\bibitem{peng2019collaborative}
Hanyu Peng, Jiaxiang Wu, Shifeng Chen, and Junzhou Huang.
\newblock Collaborative channel pruning for deep networks.
\newblock In {\em International Conference on Machine Learning}, pages
  5113--5122, 2019.

\bibitem{rastegari2016xnor}
Mohammad Rastegari, Vicente Ordonez, Joseph Redmon, and Ali Farhadi.
\newblock Xnor-net: Imagenet classification using binary convolutional neural
  networks.
\newblock In {\em European Conference on Computer Vision}, pages 525--542.
  Springer, 2016.

\bibitem{real2018regularized}
Esteban Real, Alok Aggarwal, Yanping Huang, and Quoc~V Le.
\newblock Regularized evolution for image classifier architecture search.
\newblock {\em arXiv preprint arXiv:1802.01548}, 2018.

\bibitem{ren2015faster}
Shaoqing Ren, Kaiming He, Ross Girshick, and Jian Sun.
\newblock Faster r-cnn: Towards real-time object detection with region proposal
  networks.
\newblock In {\em Advances in neural information processing systems}, pages
  91--99, 2015.

\bibitem{russakovsky2015imagenet}
Olga Russakovsky, Jia Deng, Hao Su, Jonathan Krause, Sanjeev Satheesh, Sean Ma,
  Zhiheng Huang, Andrej Karpathy, Aditya Khosla, Michael Bernstein, et~al.
\newblock Imagenet large scale visual recognition challenge.
\newblock {\em International Journal of Computer Vision}, 115(3):211--252,
  2015.

\bibitem{savva2019habitat}
Manolis Savva, Abhishek Kadian, Oleksandr Maksymets, Yili Zhao, Erik Wijmans,
  Bhavana Jain, Julian Straub, Jia Liu, Vladlen Koltun, Jitendra Malik, et~al.
\newblock Habitat: A platform for embodied ai research.
\newblock {\em arXiv preprint arXiv:1904.01201}, 2019.

\bibitem{stamoulis2018designing}
Dimitrios Stamoulis, Ting-Wu~Rudy Chin, Anand~Krishnan Prakash, Haocheng Fang,
  Sribhuvan Sajja, Mitchell Bognar, and Diana Marculescu.
\newblock Designing adaptive neural networks for energy-constrained image
  classification.
\newblock In {\em Proceedings of the International Conference on Computer-Aided
  Design}, page~23. ACM, 2018.

\bibitem{stamoulis2019single}
Dimitrios Stamoulis, Ruizhou Ding, Di Wang, Dimitrios Lymberopoulos, Bodhi
  Priyantha, Jie Liu, and Diana Marculescu.
\newblock Single-path nas: Designing hardware-efficient convnets in less than 4
  hours.
\newblock {\em arXiv preprint arXiv:1904.02877}, 2019.

\bibitem{tan2018mnasnet}
Mingxing Tan, Bo Chen, Ruoming Pang, Vijay Vasudevan, and Quoc~V Le.
\newblock Mnasnet: Platform-aware neural architecture search for mobile.
\newblock {\em arXiv preprint arXiv:1807.11626}, 2018.

\bibitem{pmlr-v97-tan19a}
Mingxing Tan and Quoc Le.
\newblock {E}fficient{N}et: Rethinking model scaling for convolutional neural
  networks.
\newblock In Kamalika Chaudhuri and Ruslan Salakhutdinov, editors, {\em
  Proceedings of the 36th International Conference on Machine Learning},
  volume~97 of {\em Proceedings of Machine Learning Research}, pages
  6105--6114, Long Beach, California, USA, 09--15 Jun 2019. PMLR.

\bibitem{tapaswi2016movieqa}
Makarand Tapaswi, Yukun Zhu, Rainer Stiefelhagen, Antonio Torralba, Raquel
  Urtasun, and Sanja Fidler.
\newblock Movieqa: Understanding stories in movies through question-answering.
\newblock In {\em Proceedings of the IEEE conference on computer vision and
  pattern recognition}, pages 4631--4640, 2016.

\bibitem{theis2018faster}
Lucas Theis, Iryna Korshunova, Alykhan Tejani, and Ferenc Husz{\'a}r.
\newblock Faster gaze prediction with dense networks and fisher pruning.
\newblock {\em arXiv preprint arXiv:1801.05787}, 2018.

\bibitem{wah2011caltech}
Catherine Wah, Steve Branson, Peter Welinder, Pietro Perona, and Serge
  Belongie.
\newblock The caltech-ucsd birds-200-2011 dataset.
\newblock 2011.

\bibitem{wang2017structured}
Huan Wang, Qiming Zhang, Yuehai Wang, and Haoji Hu.
\newblock Structured probabilistic pruning for convolutional neural network
  acceleration.
\newblock In {\em Proceedings of the British Machine Vision Conference
  ({BMVC})}, 2018.

\bibitem{wen2016learning}
Wei Wen, Chunpeng Wu, Yandan Wang, Yiran Chen, and Hai Li.
\newblock Learning structured sparsity in deep neural networks.
\newblock In {\em Advances in Neural Information Processing Systems}, pages
  2074--2082, 2016.

\bibitem{wu2018shift}
Bichen Wu, Alvin Wan, Xiangyu Yue, Peter Jin, Sicheng Zhao, Noah Golmant, Amir
  Gholaminejad, Joseph Gonzalez, and Kurt Keutzer.
\newblock Shift: A zero flop, zero parameter alternative to spatial
  convolutions.
\newblock In {\em Proceedings of the IEEE Conference on Computer Vision and
  Pattern Recognition}, pages 9127--9135, 2018.

\bibitem{yang2018netadapt}
Tien-Ju Yang, Andrew Howard, Bo Chen, Xiao Zhang, Alec Go, Vivienne Sze, and
  Hartwig Adam.
\newblock Netadapt: Platform-aware neural network adaptation for mobile
  applications.
\newblock {\em arXiv preprint arXiv:1804.03230}, 2018.

\bibitem{ye2018rethinking}
Jianbo Ye, Xin Lu, Zhe Lin, and James~Z Wang.
\newblock Rethinking the smaller-norm-less-informative assumption in channel
  pruning of convolution layers.
\newblock {\em International Conference on Learning Representation (ICLR)},
  2018.

\bibitem{Yu_2018_CVPR}
Ruichi Yu, Ang Li, Chun-Fu Chen, Jui-Hsin Lai, Vlad~I. Morariu, Xintong Han,
  Mingfei Gao, Ching-Yung Lin, and Larry~S. Davis.
\newblock Nisp: Pruning networks using neuron importance score propagation.
\newblock In {\em The IEEE Conference on Computer Vision and Pattern
  Recognition (CVPR)}, June 2018.

\bibitem{Yuan_2019_CVPR}
Xin Yuan, Liangliang Ren, Jiwen Lu, and Jie Zhou.
\newblock Enhanced bayesian compression via deep reinforcement learning.
\newblock In {\em The IEEE Conference on Computer Vision and Pattern
  Recognition (CVPR)}, June 2019.

\bibitem{zhao2019variational}
Chenglong Zhao, Bingbing Ni, Jian Zhang, Qiwei Zhao, Wenjun Zhang, and Qi Tian.
\newblock Variational convolutional neural network pruning.
\newblock In {\em Proceedings of the IEEE Conference on Computer Vision and
  Pattern Recognition}, pages 2780--2789, 2019.

\bibitem{zhao2019building}
Ritchie Zhao, Yuwei Hu, Jordan Dotzel, Christopher~De Sa, and Zhiru Zhang.
\newblock Building efficient deep neural networks with unitary group
  convolutions.
\newblock In {\em Proceedings of the IEEE Conference on Computer Vision and
  Pattern Recognition}, pages 11303--11312, 2019.

\bibitem{zhong2018target}
Yang Zhong, Vladimir Li, Ryuzo Okada, and Atsuto Maki.
\newblock Target aware network adaptation for efficient representation
  learning.
\newblock {\em arXiv preprint arXiv:1810.01104}, 2018.

\bibitem{zhou2018resource}
Yanqi Zhou, Siavash Ebrahimi, Sercan~{\"O} Ar{\i}k, Haonan Yu, Hairong Liu, and
  Greg Diamos.
\newblock Resource-efficient neural architect.
\newblock {\em arXiv preprint arXiv:1806.07912}, 2018.

\bibitem{zhou2019accelerate}
Yuefu Zhou, Ya Zhang, Yanfeng Wang, and Qi Tian.
\newblock Accelerate cnn via recursive bayesian pruning.
\newblock In {\em Proceedings of the IEEE International Conference on Computer
  Vision}, pages 3306--3315, 2019.

\bibitem{zhu2016trained}
Chenzhuo Zhu, Song Han, Huizi Mao, and William~J Dally.
\newblock Trained ternary quantization.
\newblock In {\em International Conference on Learning Representations}, 2017.

\bibitem{zhuang2018discrimination}
Zhuangwei Zhuang, Mingkui Tan, Bohan Zhuang, Jing Liu, Yong Guo, Qingyao Wu,
  Junzhou Huang, and Jinhui Zhu.
\newblock Discrimination-aware channel pruning for deep neural networks.
\newblock In {\em Advances in Neural Information Processing Systems}, pages
  883--894, 2018.

\end{thebibliography}
}

\clearpage
\appendix
\section{Optimization Interpretation of \emph{LeGR}}\label{appendix-optim}
\emph{LeGR} can be interpreted as minimizing a surrogate of a derived upper bound for the loss difference between (1) the pruned-and-fine-tuned CNN and (2) the pre-trained CNN. Concretely, we would like to solve for the filter masking binary variables $\mathbf{z}\in\{0,1\}^K$, with $K$ being the number of filters. If a filter $k$ is pruned, the corresponding mask will be zero ($z_k=0$), otherwise it will be one ($z_k=1)$. Thus, we have the following optimization problem:
\begin{align}
    \begin{split}
        \min_{\mathbf{z}}~\mathbb{L}(\mathbf{\Theta}\odot \mathbf{z} - \eta \sum_{j=1}^{\tau} \Delta \bm{w}^{(j)}\odot \mathbf{z})-\mathbb{L}(\mathbf{\Theta})\\
        \text{s.t.}~C(\mathbf{z})\le \zeta,
    \end{split}\label{eq:mask_problem}
\end{align}
where $\mathbf{\Theta}$ denotes all the filters of the CNN, $\mathbb{L}(\mathbf{\Theta}) = \frac{1}{|D|}\sum_{(x,y)\in D}~L(f(x|\mathbf{\Theta}),y)$ denotes the loss function of filters where $x$ and $y$ are the input and label, respectively. $D$ denotes the training data, $f$ is the CNN model and $L$ is the loss function for prediction (\emph{e.g.}, cross entropy loss). $\eta$ denotes the learning rate, $\tau$ denotes the number of gradient steps, $\Delta \bm{w}^{(j)}$ denotes the gradient with respect to the filter weights computed at step $j$, and $\odot$ denotes element-wise multiplication. On the constraint side, $C(\cdot)$ is the modeling function for FLOP count and $\zeta$ is the desired FLOP count constraint. By fine-tuning, we mean updating the filter weights with stochastic gradient descent (SGD) for $\tau$ steps. 

Let us assume the loss function $\mathbb{L}$ is $\Omega_l$-Lipschitz continuous for the $l$-th layer of the CNN, then the following holds:
\begin{equation}\label{eq:upper_bound_magnitude}
    \begin{split}
        & \mathbb{L}(\mathbf{\Theta}\odot \mathbf{z} - \eta \sum_{j=1}^{\tau} \Delta \bm{w}^{(j)}\odot \mathbf{z})-\mathbb{L}(\mathbf{\Theta})\\
        \le &~\mathbb{L}(\mathbf{\Theta}\odot \mathbf{z}) + \sum_{i=1}^K\Omega_{l(i)}\eta \norm{\sum_{j=1}^{\tau} \Delta \bm{w}^{(j)}_i\odot \mathbf{z}_i} - \mathbb{L}(\mathbf{\Theta})  \\
        \le & ~\sum_{i=1}^K \Omega_{l(i)} \norm{\mathbf{\Theta}_i}\mathbf{h}_i + \sum_{i=1}^K \Omega_{l(i)}^2\eta\tau\mathbf{z}_i \\
        = & ~\sum_{i=1}^K (\Omega_{l(i)} \norm{\mathbf{\Theta}_i} - \Omega_{l(i)}^2\eta\tau)\mathbf{h}_i + \Omega_{l(i)}^2\eta\tau,
    \end{split}
\end{equation}
where $l(i)$ is the layer index for the $i$-th filter, $\mathbf{h}=\mathbf{1}-\mathbf{z}$, and $\norm{\cdot}$ denotes $\ell_2$ norms. 

On the constraint side of equation~(\ref{eq:mask_problem}), let $R_{l(i)}$ be the FLOP count of layer $l(i)$ where filter $i$ resides. Analytically, the FLOP count of a layer depends linearly on the number of filters in its preceding layer:
\begin{align}\label{eq:resource}
R_{l(i)}=u_{l(i)} \norm{\{\mathbf{z}:\mathbf{z}_j~\forall j\in P(l(i)\}}_0,~u_{l(i)}\ge0,
\end{align}
where $P(l(i))$ returns a set of filter indices for the layer that precedes layer $l(i)$ and $u_{l(i)}$ is a layer-dependent positive constant. Let $\hat{R}_{l(i)}$ denote the FLOP count for layer $l(i)$ for the pre-trained network ($\mathbf{z}=\mathbf{1}$), one can see from equation (\ref{eq:resource}) that $R_{l(i)}\leq \hat{R}_{l(i)}~\forall i,\mathbf{z}$. Thus, the following holds:
\begin{align}\label{eq:upper_bound_cons}
C(\mathbf{1}-\mathbf{h})=\sum_i^K R_{l(i)} (1-\mathbf{h}_i)\leq \sum_i^K \hat{R}_{l(i)} (1-\mathbf{h}_i).
\end{align}

Based on equations (\ref{eq:upper_bound_magnitude}) and (\ref{eq:upper_bound_cons}), instead of minimizing equation (\ref{eq:mask_problem}), we minimize its upper bound in a Lagrangian form. That is,
\begin{align}\label{eq:min_lub}
    \begin{split}
        \min_{\mathbf{h}}~\sum_{i=1}^K \left(\alpha_{l(i)} \norm{ \mathbf{\Theta}_i } + \kappa_{l(i)}\right) \mathbf{h}_i,
    \end{split}
\end{align}
where $\alpha_{l(i)}=\Omega_{l(i)}$ and $\kappa_{l(i)}=\eta\tau \Omega_{l(i)}^2 - \lambda \hat{R}_{l(i)}$. To guarantee the solution will satisfy the constraint, we rank all filters by their scores $s_i=\alpha_{l(i)} \norm{ \mathbf{\Theta}_i} + \kappa_{l(i)}~\forall~i$ and threshold out the bottom ranked (small in scores) filters such that the constraint $C(\mathbf{1}-\mathbf{h})\le \zeta$ is satisfied and $\norm{\mathbf{h}}_0$ is maximized. That is, \emph{LeGR} can be viewed as learning to estimate $\bm{\alpha}$ and $\bm{\kappa}$ by assuming that better estimates of $\bm{\alpha}$-$\bm{\kappa}$ produce a better solution for the original objective (\ref{eq:mask_problem}) by solving the surrogate of the upper bound (\ref{eq:min_lub}).

\section{LeGR-DDPG}\label{sec:legrddpg}
\begin{figure}
    \centering
    \begin{subfigure}[t]{0.47\linewidth}
        \includegraphics[width=1\linewidth]{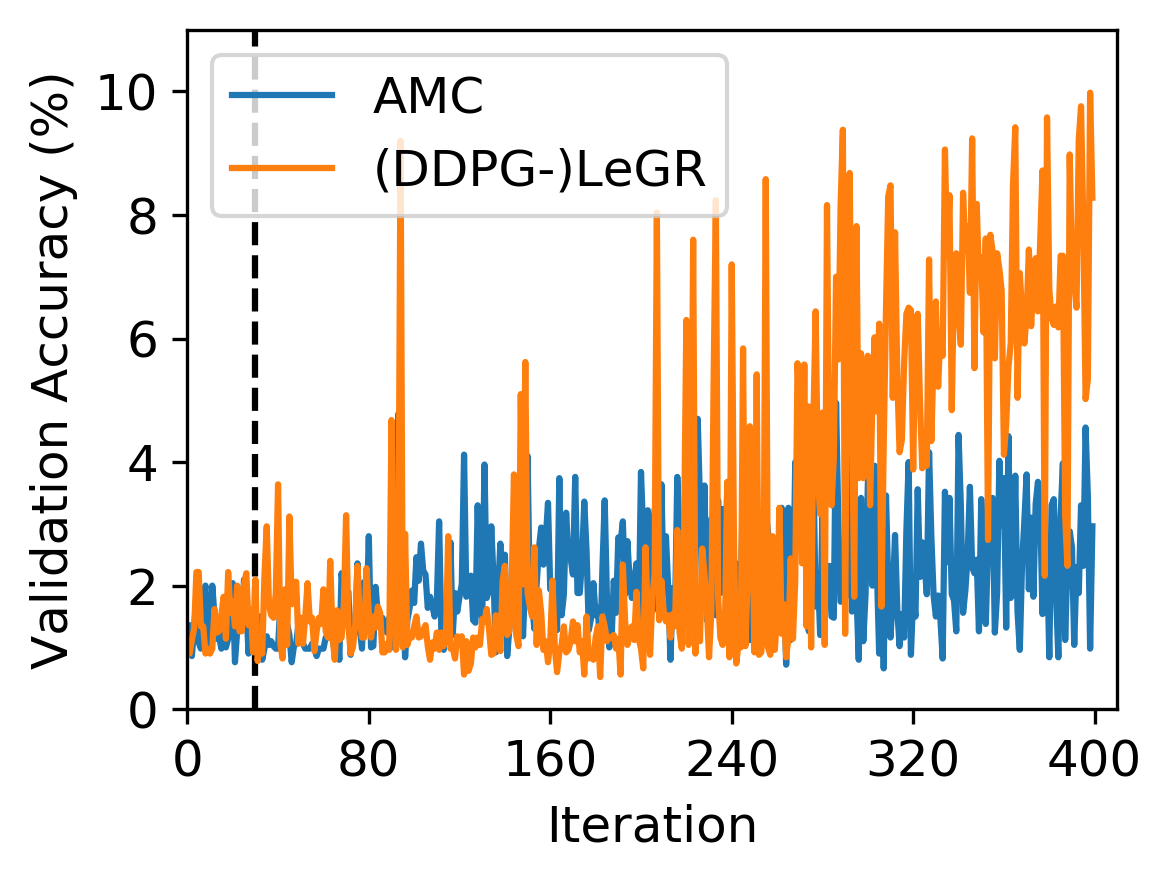}
        \caption{}
        \label{fig:rl-progress}
    \end{subfigure}
    \begin{subfigure}[t]{0.47\linewidth}
        \includegraphics[width=1\linewidth]{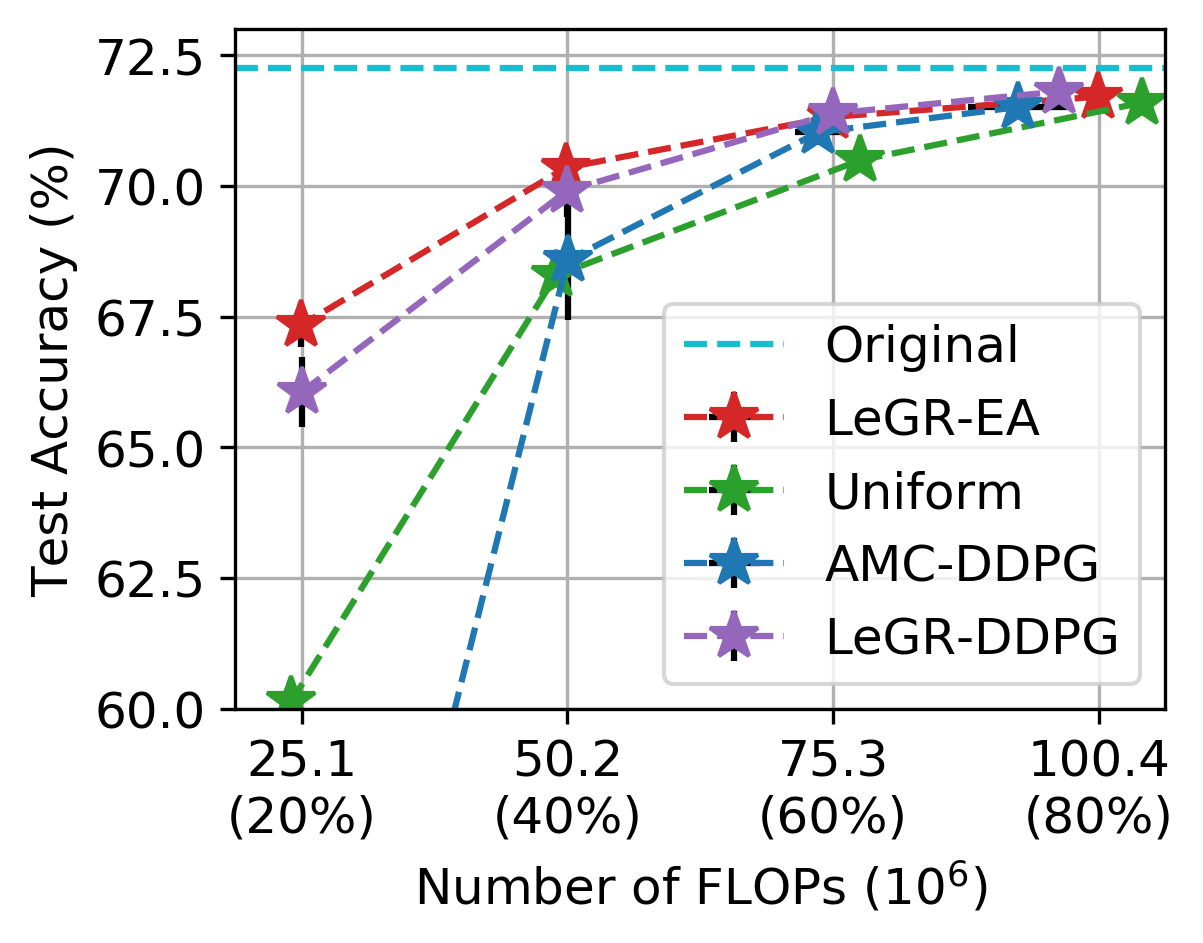}
        \caption{}
        \label{fig:norm-vs-percent}
    \end{subfigure}
    \caption{Comparison between searching the layer-wise filter norms and searching the layer-wise filter percentage. (a) compares the searching progress for 50\% FLOP count ResNet-56 and (b) compares the final performance for ResNet-56 with various constraint levels.}
\end{figure}

We have also tried learning the layer-wise affine transformations with actor-critic policy gradient (DDPG), which is adopted in prior art~\cite{he2018amc}. We use DDPG in a sequential fashion that follows~\cite{he2018amc}. \emph{LeGR} requires two continuous actions (\emph{i.e.}, $\alpha_l$ and $\kappa_l$) for layer $l$ while \emph{AMC} needs only one action (\emph{i.e.}, percentage). We conduct the comparison of pruning ResNet-56 to 50\% of its original FLOP count targeting CIFAR-100 with $\hat{\tau}=0$ and hyper-parameters following~\cite{he2018amc}. As shown in Fig.~\ref{fig:rl-progress}, while both \emph{LeGR} and \emph{AMC} outperform random search (iterations before the vertical black-dotted line), \emph{LeGR} converges faster to a better solution. Beyond comparing the progress of searching, we also compare the performance of the final pruned networks. As shown in Fig.~\ref{fig:norm-vs-percent}, searching layer-wise affine transformations is more efficient and effective compared to searching the layer-wise filter percentages. Comparing \emph{LeGR} using the two policy improvement methods, we empirically find that DDPG incurs larger variance on the final network than evolutionary algorithm.

\section{ImageNet Result Detail}\label{appendix-imagenet}
The comparison of \emph{LeGR} with prior art on ImageNet is detailed in Table~\ref{table:imagenet}. 

\begin{table*}[h]
\caption{Summary of pruning on ImageNet. The sections are defined based on the FLOP count left. The accuracy is represented in the format of \emph{pre-trained} $\mapsto$ \emph{pruned-and-fine-tuned}.}
\vskip 0.15in
\begin{center}
\begin{small}
\begin{sc}
\begin{adjustbox}{max width=0.9\textwidth}
\begin{tabular}{c|ccccccc}
\toprule
Network & Method & Top-1 & Top-1 Diff & Top-5 & Top-5 Diff & FLOP count (\%)\\
\midrule
\multirow{20}{*}{ResNet-50} &NISP~\cite{Yu_2018_CVPR} & - $\xrightarrow{}$ - & -0.2 & - $\xrightarrow{}$ - & - & 73\\
 & \textbf{LeGR} & 76.1 $\xrightarrow{}$ \textbf{76.2} & \textbf{+0.1} & 92.9 $\xrightarrow{}$ \textbf{93.0} & \textbf{+0.1} & 73\\
\cline{2-7}
&SSS~\cite{Huang_2018_ECCV} & 76.1 $\xrightarrow{}$ 74.2 & -1.9 & 92.9 $\xrightarrow{}$ 91.9 & -1.0 & 69\\
&ThiNet~\cite{luo2017thinet} & 72.9 $\xrightarrow{}$ 72.0 & -0.9 & 91.1 $\xrightarrow{}$ 90.7 & -0.4 & 63\\
& C-SGD-70~\cite{ding2019centripetal} & 75.3 $\xrightarrow{}$ 75.3 & +0.0 & 92.6 $\xrightarrow{}$ 92.5 & -0.1 & 63\\
& Variational~\cite{zhao2019variational} & 75.1 $\xrightarrow{}$ 75.2 & +0.1 & 92.8 $\xrightarrow{}$ 92.1 & -0.7 & 60\\
&GDP~\cite{lin2018accelerating} & 75.1 $\xrightarrow{}$ 72.6 & -2.5 & 92.3 $\xrightarrow{}$ 91.1 & -1.2 & 58\\
&SFP~\cite{he2018soft} & 76.2 $\xrightarrow{}$ 74.6 & -1.6 & 92.9
$\xrightarrow{}$ 92.1 & -0.8 & 58\\
&FPGM~\cite{he2019filter} & 76.2 $\xrightarrow{}$ 75.6 & -0.6 & 92.9 $\xrightarrow{}$ 92.6 & -0.3 & 58\\
&\textbf{LeGR} & 76.1 $\xrightarrow{}$ \textbf{75.7} & -0.4 & 92.9 $\xrightarrow{}$ \textbf{92.7} & -0.2 & 58\\
&GAL-0.5~\cite{lin2019towards} & 76.2 $\xrightarrow{}$ 72.0 & -4.2 & 92.9 $\xrightarrow{}$ 91.8 & -1.1 & 57\\
&AOFP-C1~\cite{ding2019approximated} & 75.3 $\xrightarrow{}$ 75.6 & \textbf{+0.3} & 92.6 $\xrightarrow{}$ \textbf{92.7} & \textbf{+0.1} & 57\\
&NISP~\cite{Yu_2018_CVPR} & - $\xrightarrow{}$ - & -0.9 & - $\xrightarrow{}$ - & - & 56\\
&Taylor-FO-BN~\cite{molchanov2019importance} & 76.2 $\xrightarrow{}$ 74.5 & -1.7 & - $\xrightarrow{}$ - & - & 55\\
\cline{2-7}
&CP~\cite{he2017channel} & - $\xrightarrow{}$ - & - & 92.2 $\xrightarrow{}$ 90.8 & -1.4 & 50\\
&SPP~\cite{wang2017structured} & - $\xrightarrow{}$ - & - & 91.2 $\xrightarrow{}$ 90.4 & -0.8 & 50\\
&\textbf{LeGR} & 76.1 $\xrightarrow{}$ \textbf{75.3} & \textbf{-0.8} & 92.9 $\xrightarrow{}$ \textbf{92.4} & -0.5 & 47\\
&CCP-AC~\cite{peng2019collaborative} & 76.2 $\xrightarrow{}$ \textbf{75.3} & -0.9 & 92.9 $\xrightarrow{}$ 92.6 & \textbf{-0.3} & 44\\
& RRBP~\cite{zhou2019accelerate} & 76.1 $\xrightarrow{}$ 73.0 & -3.0 & 92.9  $\xrightarrow{}$ 91.0 & -1.9 & 45\\
& C-SGD-50~\cite{ding2019centripetal} & 75.3 $\xrightarrow{}$ 74.5 & \textbf{-0.8} & 92.6 $\xrightarrow{}$ 92.1 & \textbf{-0.5} & 45\\
&DCP~\cite{zhuang2018discrimination} & 76.0 $\xrightarrow{}$ 74.9 & -1.1 & 92.9 $\xrightarrow{}$ 92.3 & -0.6 & 44\\
\midrule 
\multirow{6}{*}{MobileNetV2}&AMC~\cite{he2018amc} & 71.8 $\xrightarrow{}$ 70.8 & -1.0 &  $\xrightarrow{}$ \textbf{-} & \textbf{-} & 70\\
&\textbf{LeGR} & 71.8 $\xrightarrow{}$ \textbf{71.4} & \textbf{-0.4} &  $\xrightarrow{}$ \textbf{-} & \textbf{-} & 70\\
&\textbf{LeGR} & 71.8 $\xrightarrow{}$ 70.8 & -1.0 &  $\xrightarrow{}$ \textbf{-} & \textbf{-} & 60\\
\cline{2-7}
&DCP~\cite{zhuang2018discrimination} & 70.1 $\xrightarrow{}$ 64.2 & -5.9 &  $\xrightarrow{}$ - & - & 55\\
&MetaPruning~\cite{liu2019metapruning} & 72.7 $\xrightarrow{}$ 68.2 & -4.5 &  $\xrightarrow{}$ - & - & 50\\
&\textbf{LeGR} & 71.8 $\xrightarrow{}$ \textbf{69.4} & \textbf{-2.4} &  $\xrightarrow{}$ \textbf{-} & \textbf{-} & 50\\
\bottomrule
\end{tabular}\label{table:imagenet}
\end{adjustbox}
\end{sc}
\end{small}
\end{center}
\vskip -0.1in
\end{table*}

\end{document}